%% file: main.tex
\documentclass[letter,english]{article}

\usepackage[margin=1in]{geometry}

\usepackage[utf8]{inputenc} % allow utf-8 input
\usepackage[T1]{fontenc}    % use 8-bit T1 fonts

\usepackage[unicode=true,
 bookmarks=false,
 breaklinks=false,pdfborder={0 0 1},colorlinks=false]
 {hyperref}
\hypersetup{
colorlinks,citecolor=blue,filecolor=blue,linkcolor=blue,urlcolor=blue}
\usepackage{url}            % simple URL typesetting
\usepackage{booktabs}       % professional-quality tables
\usepackage{amsfonts}       % blackboard math symbols
\usepackage{nicefrac}       % compact symbols for 1/2, etc.
\usepackage{microtype}      % microtypography
\usepackage{xcolor}         % colors
\usepackage{bm}
\usepackage{amsmath}
\usepackage{amssymb}

\usepackage{multirow}
\usepackage{colortbl}
\definecolor{Gray}{gray}{0.85}
\usepackage{enumitem}

\usepackage{amsthm}
\usepackage{cite}  
\usepackage{comment}
\usepackage{natbib}
\usepackage{mathtools}

\usepackage{graphicx}
\usepackage[linesnumbered,ruled,vlined]{algorithm2e}
\usepackage{algorithmic}

\usepackage{float}
\usepackage{cancel}
\usepackage{pifont}
\usepackage{mathrsfs}
\usepackage{bbm}

\usepackage{subcaption}
\usepackage{wrapfig}

\usepackage{dsfont}
\usepackage{color}
\definecolor{yjc}{RGB}{125,0,0}
\definecolor{jiw}{RGB}{10,148,15}
\definecolor{lxs}{RGB}{138,43,226}

 \renewcommand{\hat}{\widehat}
 \renewcommand{\tilde}{\widetilde}

\input{macros}

\newcommand{\CC}{\textsf{CC}}
\newcommand{\ER}{\textsf{ER}}

\newcommand{\charter}{\textsc{Charter}}

\allowdisplaybreaks

\title{Characterizing the Accuracy-Communication-Privacy Trade-off in Distributed Stochastic Convex Optimization}
 \author{
  	Sudeep Salgia\thanks{
         Department of Electrical and Computer Engineering,
         Carnegie Mellon University;
         \texttt{\{ssalgia,yuejiec\}@andrew.cmu.edu}.}   \\
	 Carnegie Mellon   \\
	 \and
	 Nikola Pavlovic\thanks{
         Department of Electrical and Computer Engineering,
         Cornell University;
         \texttt{\{np358,qz16\}@cornell.edu}.} \\
         Cornell     \\
	 \and  
	 Yuejie Chi\footnotemark[1]\\
         Carnegie Mellon   \\
         \and
         Qing Zhao\footnotemark[2]\\
         Cornell   \\
 	} 
\date{January 2025}

\begin{document}

\theoremstyle{plain} \newtheorem{lemma}{\textbf{Lemma}}
\newtheorem{proposition}{\textbf{Proposition}}
\newtheorem{theorem}{\textbf{Theorem}}
\newtheorem{corollary}{\textbf{Corollary}}
\newtheorem{assumption}{Assumption}
\newtheorem{definition}{Definition}
\newtheorem{claim}{\textbf{Claim}}
\theoremstyle{remark}\newtheorem{remark}{\textbf{Remark}}

\maketitle

\begin{abstract}
    \input{abstract}

\end{abstract}


\setcounter{tocdepth}{2}
\tableofcontents

\input{introduction}

\input{problem_formulation}

\input{lower_bound}

\input{algorithm}

\input{discussion}

\section*{Acknowledgement}
The work of S. Salgia and Y. Chi is supported in part by the grants NSF CNS-2148212, ECCS-2318441, ONR N00014-19-1-2404 and AFRL FA8750-20-2-0504, and in part by funds from federal agency and industry partners as specified in the Resilient \& Intelligent NextG Systems (RINGS) program.

\bibliographystyle{abbrvnat}
\bibliography{references}

\newpage
\appendix

\input{proof_lower_bound}

\input{proof_upper_bound}

\end{document}

%% file: macros.tex
\newcommand{\E}{\mathbb{E}}

\newcommand{\N}{\mathbb{N}}

\newcommand{\R}{\mathbb{R}}

\newcommand{\cA}{\mathcal{A}}
\newcommand{\cB}{\mathcal{B}}

\newcommand{\cD}{\mathcal{D}}
\newcommand{\cE}{\mathcal{E}}
\newcommand{\cF}{\mathcal{F}}

\newcommand{\cJ}{\mathcal{J}}
\newcommand{\cK}{\mathcal{K}}
\newcommand{\cL}{\mathcal{L}}

\newcommand{\cN}{\mathcal{N}}
\newcommand{\cO}{\mathcal{O}}
\newcommand{\cP}{\mathcal{P}}

\newcommand{\cX}{\mathcal{X}}
\newcommand{\cY}{\mathcal{Y}}
\newcommand{\cZ}{\mathcal{Z}}

\newcommand{\bfb}{\mathbf{b}}

\newcommand{\sA}{\mathscr{A}}

\newcommand{\sM}{\mathscr{M}}

\newcommand{\sO}{\mathscr{O}}

\newcommand{\sQ}{\mathscr{Q}}

\DeclareMathOperator*{\argmin}{arg\,min}
\newcommand{\1}{\mathbbm{1}}
\newcommand{\ip}[2] {\left\langle #1,\, #2 \right\rangle }

\newcommand{\sgn}{\mathrm{sgn}}
\newcommand{\DP}{\mathsf{DP}}
\newcommand{\Err}{\mathsf{Err}}
\newcommand{\vol}{\text{vol}}

%% file: abstract.tex
We consider the problem of differentially private stochastic convex optimization (DP-SCO) in a distributed setting with $M$ clients, where each of them has a local dataset of $N$ i.i.d. data samples from an underlying data distribution. The objective is to design an algorithm to minimize a convex population loss using a collaborative effort across $M$ clients, while ensuring the privacy of the local datasets. In this work, we investigate the accuracy-communication-privacy trade-off for this problem. We establish matching converse and achievability results using a novel lower bound and a new algorithm for distributed DP-SCO based on Vaidya’s plane cutting method. Thus, our results provide a complete characterization of the accuracy-communication-privacy trade-off for DP-SCO in the distributed setting.

%% file: introduction.tex
\section{Introduction}

We consider the problem of distributed stochastic convex optimization where $M$ clients, with the aid of a central server, aim to collaboratively minimize a convex function of the form 
\begin{align}
    \cL(x) = \E_{z \sim \cP}[\ell(x;z)] \label{eqn:sco_def}
\end{align}
using their local datasets consisting of $N$ i.i.d. samples from the distribution $\cP$. Here $x \in \cX$ denotes the decision variable where $\cX$ is a convex, compact set and $\ell(x;z)$ denotes the loss at point $x$ using the datum $z$. We study this problem under the additional constraint of ensuring differential privacy~\citep{Dwork2006DPOGPaper} of the local datasets at each client. This problem arises in numerous settings and represents a typical scenario for Federated Learning (FL)~\citep{Mcmahan2017FedAvg}, which has emerged as the de facto approach for collaboratively training machine learning models using a large number of devices coordinated through a central server~\citep{Kairouz2021FLSurvey, Wang2021FieldGuideForFL}. 

Designing efficient algorithms for differentially private distributed stochastic convex optimization, also referred to as distributed DP-SCO, requires striking a careful balance between the primary objective of minimizing the optimization error and two competing desiderata --- communication cost and privacy.

\paragraph{Communication cost.} There is a natural tension between the accuracy and the communication cost of a distributed learning algorithm, as achieving a lower optimization error entails the clients sharing more information, which results in higher communication costs. Communication between the participating clients and the coordinating server is well-known to be the primary bottleneck in distributed learning, particularly in the scenario where clients have bandwidth constraints~\citep{Tang2020CommunicationSurvey, Zhao2023CommunicationSurvey}. The overall communication cost of a distributed SCO algorithm consists of two parts --- the frequency of communication and the size of the message in each communication round.
There has been a substantial effort towards designing communication-efficient algorithms in both non-private and private settings, either by reducing the frequency of communication~\citep{Mcmahan2017FedAvg, Khaled2020LocalSGDHeterogenous, Karimireddy2020Scaffold,Gorbunov2021UnifiedTheory, Li2020CommunicationVarReduction, Li2022Destress, Liu2022FedBCD, Zhao2021FedPage}, or by using compression/quantization strategies to minimize the message sizes~\citep{Konecny2016Quantization, Suresh2017DistributedMeanEstimation,Agarwal2018CPSGD,Honig2022DAdaQuant, Jhunjhunwala2021AdaptiveQuantization,Ding2021DP_Comm_Efficient_Collab_Learning, Wang2020D2P-FED, Zong2021QuantWithLocalDP, Wang2024DynamicQuantization,Li2022SoteriaFL}. 

\paragraph{Privacy.} Often in various applications, the local data at participating agents contains sensitive information that should remain private and not become publicly available during the learning process. It has been shown that preventing the transfer of the actual data during the learning process is not sufficient to guarantee privacy of the local data and can leak private information during the training process~\citep{Zhu2019DeepLeakage}. Thus, it is desirable to provide formal guarantees to protect the private data~\citep{Kairouz2021FLSurvey, Wang2021FieldGuideForFL, Geyer2017ClientPerspective}.In this work, we consider sophisticated privacy preserving techniques like differential privacy (DP)~\citep{Dwork2006DPOGPaper} to ensure the privacy of the local data.  In a seminal work,~\cite{Abadi2016DeepDP} proposed the DP-SGD algorithm where they combined SGD with DP techniques to provide formal guarantees on the privacy of the dataset for training deep networks. Since then,  numerous optimization algorithms have been proposed that ensure the privacy of the local dataset using DP. At a high level, DP ensures the privacy of the local datasets by introducing uncertainty into the output of the algorithm, which makes it difficult for an adversary to discern private information. This injection of additional uncertainty results in a natural trade-off between privacy and the accuracy of differentially private algorithms.

\paragraph{Fundamental accuracy-communication-privacy trade-off.}
Existing studies largely focus on designing algorithms that aim to balance accuracy with one of the two desiderata which provides only a partial picture of the three-way trade-off among accuracy, communication, and privacy for the problem of distributed DP-SCO. Moreover, there lack studies that characterize the converse region of this three-way trade-off, leaving the question of the optimality of existing results open. In this work, we aim to study and characterize this three-way trade-off from first principles to provide a fresh perspective and new insights into this fundamental problem.

\subsection{Main results}

We consider the problem of distributed stochastic convex optimization with $M$ clients, each with a local dataset of $N$ points, under the constraint of $(\varepsilon_{\DP}, \delta_{\DP})$ differential privacy (See Section~\ref{sec:problem_formulation} for the precise definition). In this work, we provide a complete characterization of the accuracy-communication-privacy trade-off for this problem. The accuracy of an optimization algorithm refers to the sub-optimality gap or the excess risk and is measured as $\cL(\widehat{x}) - \min_{x \in \cX} \cL(x)$, where $\widehat{x}$ denotes the output of the algorithm. We summarize the main results of our work below.
\begin{itemize}
    \item \textit{Lower bound on the accuracy-communication-privacy trade-off:} We derive a novel lower bound on the accuracy of a distributed SCO algorithm as a function of its communication cost. Specifically, we establish that the error rate of any distributed SCO algorithm is at least $\Omega\left(\sqrt{\frac{d^2}{MN \cdot \CC}} \right)$ (ignoring other terms), where $\CC$ is the communication cost of the algorithm, measured as the total number of bits transmitted by each agent on average (See Section~\ref{sec:problem_formulation} for the precise definition) and $d$ is the dimension of the decision variable. This implies that any algorithm with order-optimal accuracy incurs a communication cost of $\Omega(d^2)$ bits. This is the \emph{first} result that tightly characterizes the lower bound of communication complexity for {\em any} distributed optimization algorithm for general convex functions. When combined with existing lower bounds on the accuracy-privacy trade-off, the proposed bound characterizes the converse region of the accuracy-communication-privacy trade-off. In particular, our lower bound implies that the accuracy for any DP-SCO algorithm is at least $\Omega\left(\sqrt{\frac{d^2}{MN \cdot \min\{\CC, dN\varepsilon_{\DP}^2\}}} \right)$, where $\CC$ is the communication cost of the algorithm and $\varepsilon_{\DP}$ is the differential privacy parameter. We establish the lower bound by showing that solving a convex optimization problem is at least as hard as solving $d$ mean estimation problems. In contrast, existing lower bounds rely on the straightforward reduction of convex optimization to estimation of an unknown vector in $d$ dimensions. This is the first result that establishes that convex optimization is \emph{significantly} harder than mean estimation, tightening existing lower bounds. 
    \item \textit{Achieving the optimal accuracy-communication-privacy trade-off:} We propose a new distributed DP-SCO algorithm, called \charter, that achieves the optimal accuracy-communication-privacy trade-off as dictated by the lower bound. In particular, we show that \charter~is an $(\varepsilon_{\DP}, \delta_{\DP})$ differentially private algorithm that achieves an excess risk of $\tilde{\cO}\left(\frac{1}{\sqrt{MN}} + \frac{\sqrt{d}}{\sqrt{M}N\varepsilon_{\DP}}\right)$ and incurs a communication cost of $\tilde{\cO}(d^2)$.\footnote{Here, $\tilde{\cO}(\cdot)$ denotes the order up to logarithmic factors.} To the best of our knowledge, this is the first algorithm to achieve optimal accuracy for the problem of distributed DP-SCO. This is also the first algorithm to achieve order-optimal communication cost even in the non-private setting.
    Our proposed algorithm, \charter, departs from the family of gradient descent methods and builds upon the classical plane cutting methods~\citep{Vaidya1996PlaneCutting, Anstreicher1997Vaidya}. This paradigm shift is the key piece of the puzzle that allows us to achieve the optimal three-way trade-off, particularly along the dimension of communication complexity. The primary observation here is that the gradient descent family adopts an optimization framework that is married to the function landscape. The over-reliance on the function landscape requires more frequent communication to counter the noisy updates, particularly when the landscape is flatter. On the other hand, our plane-cutting-based method adopts a geometric perspective akin to binary search methods which allows for constant progress independent of the function landscape thereby reducing the need for frequent communication.
\end{itemize}

\subsection{Related work}

\paragraph{DP-ERM.} The problem of empirical risk minimization, or ERM for short, aims to minimize the population loss $\cL(x)$ by minimizing the sample loss function $\widehat{\cL}(x) = \frac{1}{N}\sum_{n = 1}^N \ell(x; z_n)$ for a given dataset $\cD = \{z_n\}_{n = 1}^N$. The problem of DP-ERM has been extensively studied in the centralized setting and the upper and lower bounds on the accuracy of DP-ERM are well-known~\citep{Chaudhuri2008PrivateLR, Chaudhuri2011DPERM, Bassily2014PrivateERMLowerBounds, Jain2012DPOnlineLearning, Ullman2015PrivateMultiplicateWeights, Iyengar2019DPSCO, Wang2017FasterERM}. The problem of DP-ERM has also received significant attention in the distributed setting~\citep{Li2022SoteriaFL, Phuong2022Distributed_DP, Murata2023DIFF2, Zhang2020SparseDifferentialGaussianMasking, Wang2020D2P-FED, Ding2021DP_Comm_Efficient_Collab_Learning, Triastcyn2021DPREC, Jayaraman2018DPERM-No_Distress, Huang2015DPOP}. However, solutions of ERM are known to result in poor generalization. In particular, it has been shown that solutions of ERM lead to a sub-optimal error of $\Omega(\sqrt{d/N})$ for the problem of SCO, across a large class of functions~\citep{Feldman2016gGeneralizationOfERM}. Consequently, these results are necessarily sub-optimal for the problem of DP-SCO.

\paragraph{DP-SCO.} The gap between DP-ERM and DP-SCO was first addressed in the centralized setting by~\cite{Bassily2019OptimalSCO}, where the authors propose a new algorithm that leverages the uniform stability of SGD~\citep{Bousquet2002Stability} and achieves the order-optimal accuracy of $\cO\left(\frac{1}{\sqrt{N}} + \frac{\sqrt{d}}{N\varepsilon_{\DP}}\right)$ for the problem of SCO. Since then, there have been a series of studies~\citep{Feldman2020DPSCO_Linear_Time, Han2022DPStreamingSCO, Wang2023DPSO_NonConvex, Asi2021OptimalL1Geometry, Bassily2021NewAlgorithms, Liu2024UserLevelDPSCO, Bassily2023UserSCO, Choquette2024optimal, Kulkarni2021PrivateNonSmoothSCO, Wang2020HeavyTailed, Arora2022LinearModelDP, Song2021EvadingGLMs} that further analyze the problem of DP-SCO and propose efficient algorithms with optimal performance for a wide range of scenarios in the centralized setting. However, these results do not have an analogous version for the distributed setting. In the non-private setting, the problem of distributed SCO has been extensively studied numerous algorithms have been proposed that achieve the order-optimal accuracy of $\cO\left(1/\sqrt{MN}\right)$~\citep{ Khaled2020LocalSGDHeterogenous, Woodworth2020MinibatchVsLocal, Woodworth2020LocalBetterThanMinibatch, Reisizadeh2020FEDPAQ}.  

\paragraph{Communication-efficient algorithms.} As mentioned earlier, there is an extensive line of work that focuses on designing communication-efficient algorithms both in non-private and private (DP-ERM) settings. The best-known bound on the communication complexity of distributed SCO algorithms is $\cO(d\sqrt{MN})$~\citep{Reisizadeh2020FEDPAQ, Haddadpour2021FEDCOM}.

There is a line of work that studies lower bounds on communication complexity for various distributed learning problems like mean estimation, distribution estimation, and linear bandits~\citep{Duchi2014DistributedEstimation, Braverman2016CommunicationLowerBounds, Barnes2020LowerBoundComm, Salgia2023LinearBandits}. For the problem of convex optimization,~\cite{Korhonen2021LowerBoundComm} derive a lower bound of $\Omega(d)$ by a reduction to mean estimation.~\cite{Tsitsiklis1987CommunicationLowerBound}  derive a lower bound of $\Omega(d)$ and they conjecture a lower bound of $\Omega(d^2)$. They also partially prove their conjecture for a restricted class of communication models.~\cite{Vempala2020CommunicationComplexity} also derive a lower bound of $\Omega(d^2)$ for the problem of linear regression in the non-stochastic setting where each client only has access to a partial set of features. For the gradient descent family of algorithms without acceleration, a lower bound of $\Omega(d\sqrt{MN})$ on the communication complexity was shown by~\cite{Woodworth2018GraphOracle, Huang2022LowerBoundsComm, arjevani2015communication}. Similar results for distributed learning over general networks for the gradient descent family of algorithms under the span assumption were obtained in~\cite{scaman2019optimal, scaman2017optimal}.
In this work, we establish a lower bound of $\Omega(d^2)$ for the general stochastic convex optimization that holds for \emph{all} algorithms. Our bound improves upon the best-known bound of $\Omega(d)$ and also resolves the conjecture in~\cite{Tsitsiklis1987CommunicationLowerBound}. \cite{arjevani2015communication} also derived a $\Omega(d^2)$ lower bound for the class of algorithms that perform empirical risk minimization of quadratic functions with a single round of communication, where analogous result for a more general class of algorithms that allow for multiple rounds of communication and operate over general convex functions was left as an open question. The proposed lower bound in this work also addresses this open question.

\paragraph{Notation:} The notations $f(x) = \cO(g(x))$ and $f(x) \lesssim g(x)$ both imply that the relation $f(x) \leq Cg(x)$ holds for all $x$ for some constant $C > 0$, independent of $x$. Similarly, $f(x) = \Omega(g(x))$ and $f(x) \gtrsim g(x)$ both imply that the relation $f(x) \geq cg(x)$ holds for all $x$ for some constant $c > 0$, independent of $x$. We use $\tilde{\cO}(\cdot)$ and $\tilde{\Omega}(\cdot)$ to denote the corresponding relations above up to logarithmic factors. For $n \in \N$, we use the shorthand $[n] := \{1,2,\dots, n\}$. For any event $\cE$, we use $\cE^c$ to denote its complement. For any two vectors $v, w \in \R^d$, $\ip{v}{w}$ denotes the standard inner product and $\|v\|_2 = \sqrt{\ip{v}{v}}$ denotes the $\ell_2$-norm of vector $v$.

%% file: problem_formulation.tex
\section{Problem Formulation}
\label{sec:problem_formulation}

\paragraph{Stochastic convex optimization.} We consider a distributed learning setup which consists of a single central server and $M$ clients. Each client $m \in \{1,2,\dots, M\}$ has access to a local dataset $\cD_m = \{z_{m,n}\}_{n = 1}^N \in \cZ^N$ consisting of $N$ i.i.d. data samples from a distribution $\cP_{m}$ that takes values in a set $\cZ$. The objective of the clients is to collaboratively minimize the function:
\begin{align}
\min_{x \in \cX}    \cL(x) := \frac{1}{M} \sum_{m = 1}^M \E_{z \sim \cP_m}\left[\ell(x,z )\right], \label{eqn:sco_loss_def}
\end{align}
over an input domain $\cX$ using their local datasets $\cD_m$. Here, $\cX \subset \R^d$ is a convex, compact set, and $\ell: \cX \times \cZ \to \R$ denotes the loss function of interest. Let $R := \sup\{\|x -y\|_2 \  | \ x,y \in \cX\}$ denote the diameter of the set $\cX$. Before moving forward, we outline below some definitions and assumptions that are commonly used in the SCO literature.

\begin{definition}
    A function $f$ is called $L$-Lipschitz over $\cX$ if for all $x, x' \in \cX$, $\|f(x) - f(x')\|_2 \leq L \|x - x'\|_2$.
\end{definition}

\begin{definition}
    Let $f$ be a convex function over a domain $\cX \subset \R^d$. The subgradient of $f$ at a point $x \in \cX$, denoted by $\partial f(x)$, is given by
    \begin{align*}
        \partial f(x) = \{ c \in \R^d: f(y) - f(x) \geq c^{\top} (y - x) , \; \; \ \forall y \in \cX \}.
    \end{align*}
\end{definition}

\begin{assumption}
    The population loss function $\cL(x)$ is convex. For all $x \in \cX$ and $c \in \partial \cL(x)$, $\|c\|_2 \leq 1$.
    \label{ass:pop_loss_convexity}
\end{assumption}

Let $\partial \ell(x; z)$ denote the noisy estimate of the sub-gradient at $x$ evaluated using the data point $z$. We would like to point out that we slightly abuse the notation here for ease of presentation; $\partial \ell(x; z)$ does not necessarily correspond to a subgradient of $\ell$ as it is not assumed to be a convex function.

\begin{assumption}
    For all $m \in \{1,2,\dots, M\}$ and $x \in \cX$, and $z \in \cP_m$, $\ell(x;z)$ is $\sigma_f^2$-sub-Gaussian random variable and $\partial \ell(x; z)$ is a $\sigma_g^2$-sub-Gaussian random vector such that $\E[\partial \ell(x; z)] \in \partial \cL(x)$. This implies that for all  $v \in \R^d$ with $\|v\|_2 \leq 1$ and $\lambda \in \R$, $\E[\exp(\lambda \ip{v}{\partial\ell(x; z) - \E_{z}[\partial\ell(x; z)]})] \leq \exp\left( \lambda^2 \sigma_g^2/2d\right)$ and $\E[\exp(\lambda (\ell(x; z) - \E_{z}[\ell(x; z)]))] \leq \exp( \lambda^2 \sigma_f^2/2)$. Consequently, for all $m$ and all $x \in \cX$, $\E_{z \sim \cP_m}[\|\partial\ell(x; z) - \E[\partial\ell(x; z)]\|^2] \leq \sigma_g^2$.
    \label{ass:sub_Gaussian_noise}
\end{assumption}

Assumptions~\ref{ass:pop_loss_convexity} and~\ref{ass:sub_Gaussian_noise} are imposed on the behavior of the population loss function and the distribution of the samples, which is a relatively milder requirement than on the sample loss for each data point. The assumption on the gradient norm in Assumption~\ref{ass:pop_loss_convexity} can be relaxed to any $L > 0$ using an appropriate scaling. For simplicity, we consider the case of $L = 1$. For simplicity of notation, throughout the rest of the paper, we use $\partial \cL(x)$ to denote an element of the set $\partial \cL(x)$.

\paragraph{Accuracy.} Let $\widehat{x}_{\sA}$ denote the output of an algorithm $\sA$ when run on a loss function $\cL$. The excess risk of the algorithm $\sA$ on $\cL$ is given as
\begin{align}
    \ER(\sA; \cL) & = \cL(\widehat{x}_{\sA}) - \min_{x \in \cX} \cL(x).
\end{align}
We measure the performance of an algorithm $\sA$ using $\ER(\sA)$, where
\begin{align}
    \ER(\sA) := \sup_{\cL \in  \cF} \E[\ER(\sA; \cL)]
\end{align}
denotes the worst-case expected excess risk over the functions in $\cF$, the family of convex, $1$-Lipschitz functions. Here, the expectation is taken over the randomness in the datasets $\{\cD_m\}_{m = 1}^M$ and the algorithm $\sA$. For a prescribed error $\delta_{\Err} \in (0,1) $, we analogously define $\ER(\sA; \delta_{\Err})$ which corresponds to a bound on $\sup_{\cL \in  \cF} \ER(\sA; \cL)$ that holds with probability $1 - \delta_{\Err}$ .

\paragraph{Communication cost.} We adopt the commonly used communication model where the clients can communicate only via the server. Each client can upload messages to the server which the server can broadcast to all other clients. This is commonly referred to as the blackboard model of communication~\citep{Braverman2016CommunicationLowerBounds, Barnes2020FisherLDP}. The communication cost of an algorithm $\sA$ is measured as
\begin{align}
    \CC(\sA) = \frac{1}{M} \sum_{m = 1}^M C_m(\sA),
\end{align}
where $C_m(\sA)$ denotes the number of bits uploaded by  client $m$ during a run of algorithm $\sA$. We focus only on the upload communication costs in this work, as often they are the communication bottleneck.

\paragraph{Differential privacy.} To formally define differentially private algorithms, we use the following notion of indistinguishability.
\begin{definition}\label{def:dp}
    For a given $\varepsilon > 0$ and $\delta \in (0,1)$, two distributions $P$ and $Q$ with a common support are said to be $(\varepsilon, \delta)$ indistinguishable (denoted as $P \sim_{(\varepsilon, \delta)} Q$) if the following relation holds for all events $O$ in the probability space:
    \begin{align*}
        e^{-\varepsilon} (P(O) - \delta) \leq Q(O) \leq e^{\varepsilon} P(O) + \delta.
    \end{align*}
\end{definition}

Let $\{\cD_m , \cD_m'\}_{m = 1}^M$ be a collection of pairs of {\em neighboring datasets} such that for all $m \in \{1,2,\dots, M\}$, $\cD_m$ and $\cD_m'$ differ on at most one data point. We refer to such datasets as neighboring datasets. We call an algorithm $\sA$ to be $(\varepsilon, \delta)$ differentially private \citep{Dwork2006DPOGPaper}, if for all collections of neighboring datasets $\{\cD_m , \cD_m'\}_{m = 1}^M$, we have $\sA(\{\cD_m\}_{m = 1}^M) \sim_{(\varepsilon, \delta)} \sA(\{\cD_m'\}_{m = 1}^M)$, where the probability is taken over the randomness in $\sA$.

%% file: lower_bound.tex
\section{Lower Bound}

In this section, we investigate the converse region of the accuracy-communication-privacy trade-off. The following theorem characterizes the lower bound on the worst-case accuracy (i.e., the excess population risk) of any distributed DP-SCO algorithm as a function of communication complexity and privacy guarantees.

\begin{theorem}
    \label{thm:lower_bound}
    Consider the distributed SCO problem outlined in Eqn.~\eqref{eqn:sco_loss_def} over a domain with diameter $R$, where the underlying data distributions satisfy Assumption~\ref{ass:sub_Gaussian_noise}. The excess risk of any $(\varepsilon_{\DP}, \delta_{\DP})$ differentially private algorithm $\sA$ for the problem of distributed SCO satisfies
    \begin{align*}
        \ER(\sA) & \gtrsim R \cdot
        \max\left\{ \min\left\{  \sqrt{\frac{\sigma_g^2}{ N}},  \sqrt{\frac{\sigma_g^2d^2}{MN \CC(\sA)}}, \frac{1}{\sqrt{d}} \right\}, \ \min\left\{ \sqrt{\frac{\sigma_g^2}{MN}}, \frac{1}{\sqrt{d}}\right\}, \ \frac{\sqrt{d}}{\sqrt{M} N \varepsilon_{\DP}}  \right\}.
    \end{align*}
\end{theorem}
The proof is deferred to Appendix~\ref{sec:proof_converse}. The above theorem provides a lower bound on the accuracy of any differentially private algorithm that solves the DP-SCO problem as a function of its communication cost and privacy parameter.  This is the \emph{first} information-theoretic, algorithm-independent lower bound on the accuracy-communication trade-off of a distributed SCO algorithm, both in non-private and private settings. Several comments on the theorem are in order.

\paragraph{Accuracy-Communication trade-off.} An immediate corollary of the above theorem is a lower bound on the accuracy-communication trade-off in the non-private setting, i.e., $\varepsilon_{\DP} = \infty$ with $N = \Omega(\sigma_g^2 d)$, which reads (up to scaling of $R$ and $\sigma_g^2$)
$$   \ER(\sA)  \gtrsim \max\left\{ \min\left\{  \frac{1}{ \sqrt{ N}} ,\,  \sqrt{\frac{ d^2}{MN \CC(\sA)}} \right\} ,\, \frac{1}{\sqrt{MN}}  \right\}. $$
Our lower bound also exhibits the well-known inverse relation between accuracy and communication that has been derived for other distributed learning problems~\citep{Duchi2014DistributedEstimation, Braverman2016CommunicationLowerBounds}. Moreover, it tightens the existing lower bound from $\Omega(d)$ to $\Omega(d^2)$. No algorithm $\sA$ will achieve the excess risk $   \ER(\sA)  \lesssim  \frac{1}{\sqrt{MN}}$ --- the optimal rate in the centralized setting --- unless the communication cost satisfies
$$  \CC(\sA) \gtrsim d^2.$$
This reflects our intuitive belief about the inherent hardness of general convex optimization compared to other problems like mean estimation. We would like to emphasize that this result holds only for general convex functions and not strongly convex functions. For the case of strongly convex functions, the $\Omega(d)$ bound is tight, as shown by matching upper bounds~\citep{Reisizadeh2020FEDPAQ, Haddadpour2019, Spiridonoff2020, Salgia2024CEAL}. 

\paragraph{Comparison with existing SGD-based lower bounds.} Several existing studies~\citep{arjevani2015communication, Woodworth2018GraphOracle, Woodworth2021MinMax} have tightly characterized the communication complexity of the gradient descent family of algorithms. In particular,~\cite{Woodworth2021MinMax} show that in order to achieve an accuracy of $\Theta(1/\sqrt{MN})$, SGD and accelerated SGD require $\Theta(\sqrt{MN})$ and $\Theta((MN)^{1/4})$ rounds of communication respectively, where in each round each agent transmits a $d$-dimensional vector to the server. These results are not directly comparable with those obtained above as they only hold for smooth functions, i.e., gradient is also a Lipschitz function. On the other hand, the lower bound derived in this work allows for non-smooth and even non-differentiable functions. Moreover, the lower bounds on gradient descent algorithms~\citep{Woodworth2021MinMax, arjevani2015communication}, where the bounds are derived using the optimization dynamics, only hold in the regime $d = \tilde{\Omega}((MN)^{5/4})$. The lower bound in Theorem~\ref{thm:lower_bound} is algorithm agnostic and is based on achieving statistical efficiency using information-theoretic tools. As is typical of statistical bounds, the above theorem results in non-trivial bounds in the data-rich regime, i.e., $MN \gtrsim d$. While a thoroughly fair comparison is not possible between our results and existing ones, the lower bound in Theorem~\ref{thm:lower_bound} suggests that (non-accelerated) SGD-based algorithms that are commonly used in real-world applications incur sub-optimal communication costs in the regime $\sqrt{MN} \gtrsim d$.

\paragraph{Privacy-Communication trade-off.} The above theorem suggests that up to an extent, privacy and communication work in tandem with each other, i.e., reducing communication allows to one achieve stronger privacy guarantees and higher privacy requirements allow for reduced communication costs. Such a behavior echoes a similar result obtained in~\cite{Chen2020BreakingTrilemma, Chen2024OptimalTradeOffForDME} for the case of distributed mean estimation. This trade-off between privacy and communication for DP-SCO, however, is evident only in the very high privacy regime. Specifically, note that the privacy term will be larger than the communication term only when $\varepsilon_{\DP} = \Omega(1/\sqrt{Nd})$. Consequently, in the very-high privacy regime, i.e., $\varepsilon_{\DP} = \Omega(1/\sqrt{Nd})$, the privacy requirements allow for communication costs that scale as $o(d^2)$. However, for typical use cases, i.e., $\varepsilon_{\DP} = \Theta(1)$, this part of the trade-off is not relevant. 

\paragraph{High-level proof idea.}
We establish our lower bound by considering the behavior of any algorithm on a specifically chosen convex function. In particular, we consider a function of the form $\max_{i = 1,\dots, d}\{a_i^{\top}x - b_i\}$ for appropriately chosen vectors $\{a_i\}_{i = 1}^d$ and scalars $\{b_i\}_{i = 1}^d$. Similar constructions that take the form of a maximum over linear functions have been used in previous studies to establish other lower bounds for convex optimization~\citep{Feldman2016gGeneralizationOfERM, Nemirovskij1983Book}. We establish the bound using a two-step reduction: \textit{(i)} we first show that optimizing this function is equivalent to learning at least $\Omega(d)$ vectors from the set $\{a_i\}_{i = 1}^d$; \textit{(ii)} we then show that learning these vectors is equivalent to solving $\Omega(d)$ mean estimation problems. We arrive at the final bound by combining these observations with existing bounds on the mean estimation problem~\citep{Duchi2014DistributedEstimation, Braverman2016CommunicationLowerBounds}. We would like to point out that while the current bound is derived only for $1$-Lipschitz functions, the analysis can be extended in a straightforward manner to allow for $L$-Lipschitz functions. In such a case, the privacy related term in the lower bound in Theorem~\ref{thm:lower_bound} gets scaled by a factor of $L$, while the rest remain as is.

%% file: algorithm.tex
\section{Algorithm}

In this section, we explore the achievability frontier of the accuracy-communication-privacy trade-off. One of the key challenges in designing an optimal algorithm is to bridge the existing sub-optimality gap along the communication complexity frontier. In order to address this challenge, we revisit one of the classic convex optimization approaches --- Vaidya's plane cutting method \citep{Vaidya1996PlaneCutting}. 

\paragraph{Plane cutting methods.} The philosophy of plane-cutting methods is based on the fundamental definition of convex functions. In particular, we know that the gradient of a convex function $f$ satisfies the relation $0 \geq f(x^{\star}) - f(x)\geq  \ip{\partial f(x)}{x^{\star} - x}$, where $x^{\star} \in \argmin_{x} f(x)$. This implies the gradient of a convex function allows us to construct a separating hyperplane to identify which half of the domain contains the minimizer --- a key observation exploited in plane-cutting methods. In each iteration, they obtain the gradient at a carefully chosen point, which allows them to eliminate a constant fraction of the domain. Thus in $\cO(d \log(N))$ iterations, they can arrive within a radius of $1/\sqrt{N}$ around the minimizer. 

\paragraph{Algorithm design.} The above key property of plane-cutting methods allows us to bridge the communication sub-optimality gap in distributed SCO. Specifically, we build upon the plane-cutting methods by replacing the deterministic gradients with an estimate computed by the clients. This plane-cutting-based framework allows us to transform the original SCO problem into estimating the gradient at $\tilde{\cO}(d)$ points. Note that this is precisely the reduction that characterizes the lower bound on the communication complexity, thereby resulting in an order-optimal communication complexity.

While the plane cutting approach allows us to achieve the optimal communication complexity, we lose the uniform stability of SGD based approaches which has been shown to be crucial to achieve the optimal accuracy-privacy trade-off. Thus, to address the accuracy-privacy trade-off, we carefully design our gradient estimation routine to guarantee both privacy and generalization. 

Next, we first provide an overview of the plane-cutting method used in this work followed by a description of our proposed algorithm, \charter.\footnote{The algorithm is named \charter \ because it is based on using private planes for the plane-cutting method.}

\subsection{Vaidya's Plane Cutting Method}

Vaidya's Plane Cutting method is a classical convex optimization algorithm proposed by~\cite{Vaidya1996PlaneCutting} to minimize a convex function $f(x)$ over a given convex, compact set $\cK$. We first introduce some notation and then provide a general description of the algorithm. 

Let $P = \{x \in \R^d : Ax \geq b\}$ be a bounded $d$-dimensional polyhedron, where $A \in \R^{p \times d}$ and $b \in \R^p$. The volumetric barrier of the set $P$ is defined as 
\begin{align*}
    V(x) & := \frac{1}{2} \log(\det(H(x))), \qquad  
    \text{ where } H(x)   = \sum_{i = 1}^p \frac{a_i a_i^{\top}}{(a_i^{\top}x - b_i)^2}.
\end{align*}
Here, $a_i^{\top}$ is the $i^{\text{th}}$ row of the matrix $A$ and $\det(B)$ denotes the determinant of the matrix $B$. The minimizer of the function $V(x)$ in the interior of $P$ is referred to as the volumetric center of the set $P$. Lastly, for all $i \in \{1,2,\dots, p\}$, we define
\begin{align*}
    \sigma_i(x) := \frac{a_i^{\top}(H(x))^{-1}a_i}{(a_i^{\top}x - b_i)^2}.
\end{align*}

Vaidya's method proceeds by generating a sequence of pairs $(A_k, b_k) \in \R^{p_k \times d} \times \R^{p_k}$ such that the corresponding polyhedrons contain the solution of the problem, i.e., minimizer of the function $f$.  Here $p_k$ denotes the number of constraints used to describe the polyhedron constructed during the $k^{\text{th}}$ iteration. The initial polyhedron $(A_0, b_0)$ is taken to be a unit hypercube. For simplicity, we assume that $\cK$ corresponds to this hypercube. The algorithm can be easily modified to the case where $(A_0, b_0)$ is a bounding hypercube of the set $\cK$. Vaidya's method also uses two hyperparameters $\eta, \gamma \in (0,1)$ which are numerical constants independent of problem parameters. The parameter $\gamma$ is used to control the total number of constraints in any given iteration by eliminating constraints that are less important. The parameter $\eta$, together with $\gamma$, determines the rate of progress in each iteration.

At the beginning of each iteration $k \geq 0$, the learner determines the approximate volumetric center $x_k$ and calculates $\{\sigma_i(x_k)\}_{i = 1}^{p_k}$. The next polyhedron characterized by the pair $(A_{k+1}, b_{k+1})$ is obtained from the current result by either adding or removing a constraint. In particular,
\begin{itemize}
    \item if, $\sigma_{i}(x_k) = \min_{1\leq j\leq p_k} \sigma_j(x_k) < \gamma$, then $(A_{k+1}, b_{k+1})$ is obtained by eliminating the $i^{\text{th}}$ row from $(A_k, b_k)$;
    \item otherwise, the algorithm first determines a $\beta_k \in \R$, such that %\yc{should $b_k$ be $\beta_k$ in the equation below?}
    \begin{align*}
        \frac{c_k^{\top}(H(x_k))^{-1}c_k}{(c_k^{\top}x - \beta_k)^2} = \frac{1}{2}\sqrt{\eta \gamma},
    \end{align*}
    where $c_k \in -\partial f(x_k)$ (which is the subgradient of $f$ at $x_k$) and then adds the constraint $(c_k^{\top}, \beta_k)$ to $(A_k, b_k)$ to obtain $(A_{k+1}, b_{k+1})$.
\end{itemize}

Vaidya's method has been studied in great detail since it was proposed by Vaidya. We refer the reader to~\cite{Anstreicher1997Vaidya, Anstreicher1998PracticalVaidya, Ye1996Complexity, Lee2015Faster, Jiang2020OptimalCuttingPlaneComplexity} and references therein for additional details about the implementation and hyperparameter choices. Vaidya's method has also been studied for non-private, stochastic convex optimzation in the centralized setting~\citep{Feldman2021SQAlgorithms, Gladin2021MinMax, Gladin2022vaidya, Mehrotra2000VolumetricSampling}. We extend results in these studies to a distributed setting with differential privacy. Also, \charter \ offers an improved statistical complexity over these existing studies.

\subsection{The \charter \ algorithm}

The proposed algorithm builds upon the classical plane cutting methods, while incorporating the elements of stochasticity and privacy. The algorithm consists of two stages, the learning stage and the verification stage. Before the start of the algorithm, each client splits their dataset into two parts, namely $\cD_m^{(1)}$ and $\cD_m^{(2)}$ consisting of $2N/3$ and $N/3$ samples respectively.\footnote{Without loss of generality, we assume $N$ is divisible by $3$.}

\paragraph{The learning stage.} The first stage of the algorithm generates a sequence of iterates $\{x_0,x_1, \dots, x_K\}$ using $K$ iterations of the Vaidya's method, where $K$ is a parameter of the algorithm. In the $k^{\text{th}}$ iteration, the cutting plane is constructed using an estimate of $\partial \cL(x_{k-1})$ which is computed collaboratively by the clients.

In order to collaboratively estimate the gradient at a given point $x$, each client $m$ first computes 
\begin{align}
    \widehat{\partial \cL}_m^{\textsf{NonPriv,b}}(x)  & 
 :=   \frac{3K}{N} \sum_{z \in \cD_m^{(1,k)}} \textsf{clip}(\partial \ell(x; z);  G_0) \cdot 
     \1\{z \notin \cup_{j = 1}^{k-1} \cD_m^{(1, j)}\}, \label{eqn:non_private_biased_estimate}
\end{align}
which is an estimate of $\partial \cL(x)$ based on the local data at the client $m$. Here $\cD_m^{(1,k)}$ is a subset of size $N/3K$\footnote{The batch size can be set to $\lceil N/3K \rceil$ to ensure it is an integer. We ignore the divisibility issue for the ease of presentation.} drawn randomly from the set $\cD^{(1)}_m$ during the $k^{\text{th}}$ iteration. The \textsf{clip} is the standard clipping function, where \textsf{clip}$(y, G) = y \cdot \min\{1, G/\|y\|_2\}$. Note that, in order to ensure that the estimated gradient is an independent sample of  $\partial \cL(x)$, we only use the samples that have not been seen before, as denoted by the indicator function $\1\{\cdot\}$. This is crucial to guarantee generalization. This, however, introduces a bias in the estimate (denoted by the superscript $\textsf{b}$), which we correct for in a later step. The non-private estimate is then privatized using the Gaussian mechanism to obtain,
\begin{align}
    \widehat{\partial \cL}_m^{\textsf{Priv,b}}(x) := \widehat{\partial \cL}_m^{\textsf{NonPriv,b}}(x) + \cN(0, \sigma_0^2 I_d). \label{eqn:private_biased_estimate}
\end{align}
After privatizing, we debias the gradient estimate by dividing the privatized estimate by $3KT_{k, m}/N$, to obtain
\begin{align}
    \widehat{\partial \cL}_m^{\textsf{Priv,u}}(x) := \frac{N}{3K T_{k, m}} \cdot \widehat{\partial \cL}_m^{\textsf{Priv,b}}(x). \label{eqn:private_unbiased_estimate}
\end{align}
Here $T_{k,m} = |\cD_m^{(1,k)} \setminus \bigcup_{j = 1}^{k-1} \cD_m^{(1,j)}|$ is the number of unseen elements in $\cD_m^{(1,k)}$. We use this two step procedure to estimate the gradient to ensure both \emph{privacy} and \emph{generalization}. Specifically, in order to ensure generalization, we need to ensure that we use an independent estimate of $\partial \cL(x)$ in each iteration. A straightforward way to guarantee that is to randomly sample from the subset of samples not seen so far. However, this does not allow us leverage privacy amplification guarantees through subsampling as the randomness in the algorithm becomes dependent across different calls to the dataset. Thus, to address this issue we always sample from the entire dataset, which helps us obtain optimal privacy dependence through subsampling and composition~\citep{Balle2018PrivacyAmplification, Steinke2022Composition, Dwork2015AdaptiveComposition, Kairouz2015Composition}. To obtain generalization, we drop the previously seen samples in constructing our estimate to ensure independence of the samples. We carefully choose our batch sizes to ensure that $3KT_{k, m}/N = \Theta(1)$ holds with high probability for all iterations so that the utility of the algorithm worsens by no more than a constant factor when we debias the gradient after privatization. We would like to point out that $T_{k,m}$ is independent of the \emph{actual} value of the samples. As a result, the debiasing step is effectively a post-processing step and thus maintains the privacy of the estimate. Lastly, we quantize the privatized estimate to obtain 
\begin{align}
    \widehat{\partial \cL}_m(x) := \sQ(\widehat{\partial \cL}_m^{\textsf{Priv,u}}(x); D_0, J_0). \label{eqn:quantized_estimate}
\end{align}
Here, $\sQ$ is the standard stochastic quantization routine~\citep{Suresh2017DistributedMeanEstimation} that separately clips each coordinate to within the interval $[-D_0, D_0]$ and quantizes it using $J_0$ bits. Specifically, the quantizer first splits the interval $[-D_0, D_0]$ into $2^{J_0} -1$ intervals of equal length where $-D_0 = r_1 < r_2 \ldots < r_{2^{J_0}} = D_0$ correspond to end points of the intervals. Each coordinate of input vector $w$ is then separately quantized as follows. The value of the $p$-th coordinate, $\sQ(w)[p]$,  is set to be $r_{j_p - 1}$ with probability $\frac{r_{j_p} - w[p]}{r_{j_p} - r_{j_p -1}}$ and to $r_{j_p}$ with the remaining probability, where $j_p := \min\{j : r_j < w[p] \leq r_{j+1}\}$. It is straightforward to note that each coordinate of $\sQ(w)$ can be represented using $J_0$ bits and has an error of at most $2D_0 \cdot 2^{-J_0}$.

Finally, each client transmits the quantized version $\widehat{\partial \cL}_m(x)$ to the server, where it evaluates 
\begin{align}
    \widehat{\partial \cL}(x) = \frac{1}{M} \sum_{m = 1}^M \widehat{\partial \cL}_m(x)
\end{align}
and sends it back to the clients to be used in the Vaidya's plane cutting method.

\paragraph{The verification stage.} In the second stage, each client uses their local dataset $\cD^{(2)}_m$ to estimate the value of $\cL(x)$ for all the $K+1$ iterates, $\{x_0, x_1, \dots, x_K\}$, generated during the learning stage. The values are computed using a similar three step procedure as used in the learning stage, i.e., estimation, privatization and quantization. In particular, for each $x \in \{x_0, x_1, \dots, x_K\}$, each agent computes
\begin{align}
    \widehat{\cL}^{\textsf{NonPriv}}_m(x) & := \frac{3}{N} \sum_{z \in \cD_m^{(2)}} \ell(x; z) \cdot \1\{|\ell(x; z)| \leq G_1\}, \label{eqn:loss_estimate} \\
    \widehat{\cL}^{\textsf{Priv}}_m(x) & := \widehat{\cL}^{\textsf{NonPriv}}_m(x) + \cN(0, \sigma_1^2), \label{eqn:loss_privatize} \\
    \widehat{\cL}_m(x) & := \sQ(\widehat{\cL}^{\textsf{Priv}}_m(x); D_1, J_1). \label{eqn:loss_quantize}
\end{align}

The local estimates $\{\widehat{\cL}_m(x_k)\}_{k = 0}^K$ are sent to the server, where they are averaged, and the index 
\begin{align}
    k^{\star} := \argmin_{k} \frac{1}{M} \sum_{m = 1}^M \widehat{\cL}_m(x_k) \label{eqn:k_star_def}
\end{align}
is returned by the server. The output of the algorithm is set to $x_{k^{\star}}$. A pseudocode of the algorithm is presented in Algorithm~\ref{alg:algorithm}.

\input{pseudocode}

\subsection{Setting the parameters}
\label{sub:parameter_setting}

The desired performance of the algorithm is obtained by carefully choosing the parameters in both stages. Let $\varepsilon_{\DP} > 0$ and $\delta_{\DP} \in (0,1)$ denote the privacy parameters and let $\delta_{\Err} \in  (0,1)$. We follow the following choices of parameters. 
\begin{itemize}
\item The number of iterations is set to $K := \left\lceil (4d/\gamma) \log\left(\frac{d\sqrt{MN}}{\gamma\sigma_g}\right) \right\rceil $, where $\gamma$ is the parameter of Vaidya's method. 
\item The clipping radii are set to $G_0 := 1 + \sigma_g\sqrt{2 \log(4MN)}$ and $G_1 := R + \sigma_f\sqrt{2 \log(4MN)}$. 
\item The privacy noise parameters are set to $\sigma_0^2 := \frac{1080 G_0^2 \log^2(2.5/\delta_{\DP}) K}{N^2 \varepsilon_{\DP}^2}$ and $\sigma_1^2 := \frac{40G_1^2 \log^2(2.5K/\delta_{\DP})K}{N^2 \varepsilon_{\DP}^2}$. \item The quantization parameters are set to $D_0 := G_0 + \sigma_0 \sqrt{32 \log\big(\frac{40MKd}{\delta_{\Err}} \big)}$, $D_1 := G_1 + \sigma_1 \sqrt{2 \log\big(\frac{16MK}{\delta_{\Err}} \big)} $, $J_0 := \left \lceil \log_2\big(\frac{2D_0N\varepsilon_{\DP}}{\sqrt{d} + \sigma_g\varepsilon_{\DP}\sqrt{N}} \big) \right\rceil$ and $J_1 := \left\lceil \log_2\big(\frac{2D_1N\varepsilon_{\DP}}{R\sqrt{d} + \sigma_f\varepsilon_{\DP}\sqrt{N}} \big) \right\rceil$.
\end{itemize}

\subsection{Performance guarantees}

The following theorem characterizes the performance of the proposed algorithm.

\begin{theorem}
    Assume that Assumptions~\ref{ass:pop_loss_convexity} and~\ref{ass:sub_Gaussian_noise} hold and the domain $\cX$ is a hypercube. If \charter \ is run with the choice of parameters described in Section~\ref{sub:parameter_setting} with $N = \Omega(d \log(KM))$ samples at each agent, then for any given privacy parameters $\varepsilon_{\DP} \in (0, 1.5/\sqrt{K})$ and $\delta_{\DP} \in (0,1)$, and error probability $\delta_{\Err} \in (0,1)$, 
    \begin{itemize}
        \item \charter \ is $(\varepsilon_{\DP}, \delta_{\DP})$ differentially private;
        \item The error rate of \charter \ satisfies %the with probability $1 - \delta_{\Err}$
        \begin{align*}
            \ER( \charter \ ; \delta_{\Err}) = \tilde{\cO} \left( \frac{R\sigma_g + \sigma_f}{\sqrt{MN}} + (R(1 + \sigma_{g}) + \sigma_{f}) \cdot \frac{\sqrt{d}}{N\varepsilon_{\DP} \sqrt{M}} \right);
        \end{align*}
        \item The communication cost of \charter \ satisfies
        \begin{align*}
            \CC(\charter) = Kd (J_0 + J_1) = \tilde{\cO}(d^2).
        \end{align*}
    \end{itemize}
    \label{thm:upper_bound}
 \end{theorem}

 A proof of the above theorem can be found in Appendix~\ref{sec:proof_achievability}. For the case of general $L$-Lipschitz functions, the term $(1+\sigma_g)$ gets updated to $(L + \sigma_g)$. A few implications of the theorem are in order.

 \paragraph{Optimal Accuracy-Communication-Privacy Trade-off.} As shown by the above theorem, \charter \ is differentially private, achieves the optimal accuracy, including linear speedup w.r.t. the number of clients, and order-optimal communication complexity (for $\varepsilon_{\DP} \geq \sqrt{d/N}$) as dictated by the lower bound derived in the previous section. Thus, \charter \ is the \emph{first} algorithm to achieve order-optimal performance on all the three fronts for distributed, differentially private stochastic optimization of general convex functions. Together with our lower bound, it provides tight characterization of the frontier for $\varepsilon_{\DP} \leq \frac{1.5}{\sqrt{K}}$. This constraint on the privacy parameter is a consequence of using privacy amplification by subsampling without replacement which holds only for $\varepsilon_{\DP} < 1$~\citep{Balle2018PrivacyAmplification},. We believe this can be resolved by utilizing a different privacy amplification scheme. We leave the extension to future work. We would also like to point out that we assume $\cX$ to be a hypercube only for convenience. The result extends immediately to general convex bodies by appropriately incorporating the change in Vaidya's method.

 \paragraph{Beyond the three-way trade-off.} In addition to achieving optimal performance along all the three desiderata, \charter \ also possesses several other desirable properties. Theorem~\ref{thm:upper_bound} holds for general, convex, Lipschitz functions without any assumption on smoothness of the function, as is required by numerous existing studies. Moreover, in terms of gradient complexity, \charter \ requires only $\cO(N)$ gradient computations,\footnote{We refer to the computation of the gradient at a single data point as a unit computation. For the function evaluation, we compute $K = \tilde{\cO}(d)$ scalar values, which is computationally equivalent to $\tilde{\cO}(1)$ gradient evaluations.} which improves upon the current state of the art for distributed algorithms~\citep{Murata2023DIFF2} and matches that in the single agent setting~\citep{Choquette2024optimal}. Furthermore, note that we do not require the data distribution to be identical for all the clients. For each point $x_k$, we use an unbiased estimate of the gradient based on data from \emph{all} the clients. As a result, the gradient estimated at the server is always an unbiased estimate of the true gradient of $\cL(x)$, even when the client distributions are different. Thus, \charter \ also lends itself to scenarios with heterogeneous data distribution. Lastly, \charter \ also allows seamless integration with client sampling. In particular, if at each communication round only a $s \in (0,1)$ fraction of clients are available, \charter \ offers a similar error rate guarantee with $M$ replaced with $sM$.

%% file: pseudocode.tex
\begin{algorithm}[h]
    \caption{\charter: At client $m$}
    \label{alg:algorithm}
    \begin{algorithmic}[1]
        \STATE Input: Initial point $x_0$
        \STATE Divide the local dataset into $\cD^{(1)}_m$ and $\cD^{(2)}_m$
        \STATE \texttt{// Set the parameters as described in Sec.~\ref{sub:parameter_setting}}
        \STATE \texttt{ //Learning Stage}
        \FOR{$k = 0,1,\dots, K$}
        \STATE Sample a subset $\cD_{m}^{(1, k)}$ of size $N/3K$ uniformly at random from $\cD_m$
        \STATE Compute the estimate $\widehat{\partial \cL}_m^{\textsf{NonPriv,b}}(x_k)$ using Eqn.~\eqref{eqn:non_private_biased_estimate} 
        \STATE Compute $\widehat{\partial \cL}_m^{\textsf{Priv,b}}(x_k)$ using Eqn.~\eqref{eqn:private_biased_estimate}
        \STATE Compute $\widehat{\partial \cL}_m^{\textsf{Priv,u}}(x_k)$ using Eqn.~\eqref{eqn:private_unbiased_estimate}
        \STATE Quantize the current estimate using Eqn.~\eqref{eqn:quantized_estimate} to obtain $\widehat{\partial \cL}_m(x_k)$
        \STATE Transmit $\widehat{\partial \cL}_m(x_k)$ to the server and receive $\widehat{\partial \cL}(x_k)$
        \STATE Use Vaidya's Method with $\widehat{\partial \cL}(x_k)$ to compute $x_{k+1}$
        \ENDFOR
        \STATE \texttt{ // Verification Stage}
        \FOR{$k = 0,1,\dots, K$}
        \STATE Evaluate $\widehat{\cL}_m(x_k)$ using Eqns.~\eqref{eqn:loss_estimate},~\eqref{eqn:loss_privatize} and~\eqref{eqn:loss_quantize}
        \ENDFOR
        \STATE Transmit $\{\widehat{\cL}_m(x_k)\}_{k =0}^{K}$ to the server and receive $k^{\star}$
        \STATE \textbf{return} $x_{k^{\star}}$
    \end{algorithmic}
\end{algorithm}

%% file: discussion.tex
\section{Discussions and Future Work}

Our approach presents a departure from the existing SGD family of algorithms and adopts a different philosophical outlook toward the optimization problem. Specifically, SGD adopts a ``function landscape'' based optimization approach, where it moves down the the function landscape until it reaches the bottom of the valley, or equivalently the minimum value. On the other hand, \charter \ adopts a more geometric perspective to the optimization problem where the objective is to eliminate regions of the domain that do not contain the minimizer, reminiscent of the classic bisection algorithms in one dimension~\citep{Frazier2019PBA, Vakili2019RWT}. While both approaches offer similar accuracy and privacy performances, the key difference is reflected in their communication complexities. The ``function landscape'' based approach is inherently tied to the steepness of the function valley. When the valley is wide, over reliance on the local steps taken by the agents precludes the algorithm from determining a useful descent direction, resulting in a bias (also referred to as the client drift) that leads to a sub-optimal performance. In order to remedy this and prevent excessive reliance on local steps, SGD-based algorithms need to communicate frequently, which results in high communication complexity. On the other hand, the geometric perspective to optimization avoids this pitfall and can eliminate sub-optimal regions at a constant rate, thereby requiring less frequent communication. A similar conclusion in the context of adapting to function regularity was noted in~\cite{Vakili2019RWT}. 

While this geometric perspective offers an improved communication complexity, it comes at the cost of increased computation complexity. Specifically, Vaidya's method has a computation complexity of $\cO(d^3 + \textsf{grad})$, where $\textsf{grad}$ denotes the overall computational complexity of evaluating all the gradients~\citep{Jiang2020OptimalCuttingPlaneComplexity}. This order of scaling prevents the application of our proposed approach to high-dimensional problems. Given the inherent necessity of adopting a geometric perspective to achieve optimal communication complexity, this suggests that the three-way trade-off is likely a four-way trade-off with computational complexity as the fourth axis. An interesting future direction is to explore if and how computational complexity poses a bottleneck in achieving the optimal accuracy-communication-privacy trade-off.

Another interesting direction is to iron out the small sub-optimality region for the communication cost in the high-privacy regime. We believe this can be remedied using more sophisticated quantization schemes that combine privacy and quantization~\citep{Chen2024OptimalTradeOffForDME}. The proposed scheme in~\cite{Chen2024OptimalTradeOffForDME} uses at most a single bit for each dimension. However, in our setting it might be necessary to use multiple bits to represent the values in each dimension which precludes a direct adaptation of that approach. Designing quantization schemes that allow for multiple-bit representation while ensuring privacy is another direction worth exploring.

%% file: proof_lower_bound.tex
\section{Proof of Theorem~\ref{thm:lower_bound}}
\label{sec:proof_converse}
In order to establish the lower bound, we focus on bounding the accuracy as a function of the communication cost without the constraint on privacy. The proof consists of three main steps.

\begin{itemize}
    \item Constructing the ``hard'' instance: We first construct a function of interest on which we analyze the performance of a distributed SCO algorithm. This is a typical step in establishing lower bounds, where the function of interest is chosen to reflect the inherent hardness of the problem.
    \item Reduction to mean estimation:  In the second step, we show that optimizing the above function is at least as hard as solving $\Omega(d)$ mean estimation problems. This step allows us to reduce the original convex optimization problem to a set of simpler problems for which we understand the accuracy communication trade-off.
    \item Establishing the final bound: The final bound is then established by combining the above reduction with techniques developed for the mean estimation problem.
\end{itemize}

\subsection{Constructing the instance of interest}

Let $\{a_1, a_2, \dots, a_d\}$ be any orthonormal basis of $\R^d$. Let $\bfb = (b_1, b_2, \dots, b_d) \in \{-1,1\}^d$. Throughout the proof, we set $\cX = \cB(1)$, the unit ball in $\R^d$ centered at the origin. For all $i \in \{1,2,\dots, d\}$, we define the following function:
\begin{align*}
    f_i(x) = \left| a_i^{\top}x - \frac{b_i}{\sqrt{d}}\right|.
\end{align*}
The results in the lower bound can be extended immediately to a ball of radius $R/2$ by replacing $b_i$ with $Rb_i/2$ throughout the proof. For simplicity of notation we present the proof with $R = 2$.

We will consider the following objective function for the analysis:
\begin{align}
    f(x) = \alpha \cdot \max_{i = 1,2, \dots, d} f_i(x),
\end{align}
where $\alpha \in (0,1]$ is a parameter, whose value is chosen later. Note that $\{f_i(x)\}_{i = 1}^d$ is a collection of convex, $1$-Lipschitz functions. Since taking the maximum operation preserves this property, $f(x)$ is also a convex, $1$-Lipschitz function. Let $\cF'$ denote the class of functions of the above form corresponding to different choices of the orthonormal basis $\{a_1, a_2, \dots, a_d\}$ and the vector
$\bfb$.

It is straightforward to note that $f(x) \geq 0$ for all $x \in \cX$ and $f(x)$ has a unique minimizer $x^{\star}$ with $f(x^{\star}) = 0$ where 
\begin{align}
    x^{\star} := \frac{1}{\sqrt{d}}\sum_{i = 1}^d a_i b_i.
\end{align}

We consider the following model for the subgradient observations. In particular, the subgradient of any randomly drawn data point $z$ is distributed as 
\begin{align}
    \partial f(x;z) \sim \alpha \cdot a_{i(x)} \cdot s_{i(x)}(x) + \cN(0, (\sigma^2/d) \cdot I_d), \label{eqn:original_noisy_oracle}
\end{align}
where for all $j \in \{1,2,\dots, d\}$, $s_j(x)$ is defined as:
\begin{align}
    s_j(x) := \begin{cases} +1 \text{ if } a_j^{\top}x - \frac{b_j}{\sqrt{d}} \geq 0, \\ -1 \text{ otherwise},\end{cases}
\end{align}
and $i(x)$ is given by:
\begin{align}
    i(x) := \min \ \{j \in \{1,2,\dots, d\} \ | \ f(x) = f_j(x)\}.
\end{align}
In other words, $s_j(x)$ determines the sign of the $a_j^{\top}x - \frac{b_j}{\sqrt{d}}$ and $i(x)$ is the smallest index from the set of the functions that achieve the maximum value at $x$. It is straightforward to note that this distribution satisfies Assumption~\ref{ass:sub_Gaussian_noise} with $\sigma_g = \sigma$. Let $\cA$ denote the class of algorithms that can optimize functions in $\cF'$ using observations of the form~\eqref{eqn:original_noisy_oracle}.

For analytical convenience, we adopt the framework of having an oracle for the noisy gradients. Specifically, the algorithm is allowed $N$ queries to an oracle $\sO$ at each agent. Each query to the oracle $\sO$ reveals a noisy gradient at the queried point $x$ that follows the same distribution as in~\eqref{eqn:original_noisy_oracle}. Thus, querying an oracle is equivalent to computing the gradient at a new data point. 

The benefit of the oracle framework is that it allows us to consider a more powerful oracle for the subgradient. Specifically, we consider an oracle $\sO'$ which when queried at a point $x$ reveals the tuple
\begin{align*}
    \sO'(x) = (\alpha \cdot a_{i(x)} + \cN(0, (\sigma^2/d) \cdot I_d), s_{i(x)}(x)),
\end{align*}
i.e., it separately provides the gradient and the sign information. Clearly, $\sO'$ is a more informative oracle. Consequently, if $\cA$ and $\cA'$ denote the class of algorithms that can optimize functions in $\cF'$ using observations from oracles $\sO$ and $\sO'$ respectively, then $\cA \subseteq \cA'$. For the remainder of the analysis, we focus on the algorithms in the class $\cA'$.

\subsection{Reduction to mean estimation}

In this part, we show that any algorithm $\sA \in \cA'$ that achieves a small optimization error, needs to solve at least $\Omega(d)$ mean estimation problems. We establish this reduction in four steps.

\paragraph{Step 1: Low optimization error is equivalent to learning $\bfb$.} For any $A = [a_1, a_2, \dots, a_d]$ and $\bfb$, a given algorithm $\sA$ and all $j \in \{1,2,\dots, d\}$, define
\begin{align}
    p_j(\sA; A,\bfb) = \Pr \left( \sgn(a_j^{\top} \widehat{x}_{\sA}) = b_j \right),
\end{align}
where $\widehat{x}_{\sA}$ denotes the output of $\sA$ when run on the function corresponding to $(A, \bfb)$ and $\sgn(\cdot)$ is the sign function. Here the probability is taken over the randomness in $\sA$ and noise distribution. If $(A', \bfb')$ is an instance such that for some index $i$, $p_i(\sA; A',\bfb') < 5/6$, then for the function $f$ corresponding to $(A', \bfb')$,
\begin{align*}
    \E[\ER(\sA; f)] = \E[f(\widehat{x}_{\sA})] & \geq \E[\alpha f_i(\widehat{x}_{\sA})] \\
    & \geq \E\left[ \alpha \left|a_i^{\top} \widehat{x}_{\sA} - \frac{b_i}{\sqrt{d}}\right| \ \bigg| \ \sgn(a_i^{\top} \widehat{x}_{\sA}) \neq b_i \right] \cdot \Pr(\sgn(a_i^{\top} \hat{x}_{\sA}) \neq b_i)  > \frac{\alpha}{6\sqrt{d}}.
\end{align*}
For the first equality we use the fact that the minimum value of $f$ is $0$. Thus, for any algorithm $\sA$ 
\begin{align}
    \sup_{f \in \cF'} \E[\ER(\sA, f)] \leq \frac{\alpha}{6\sqrt{d}} \implies \max_{j} \sup_{(A,\bfb)} p_j(\sA; A,\bfb) \geq \frac{5}{6}. \label{eqn:low_error_implies_learning_b}
\end{align}
In other words, if $\sA$ achieves a small excess risk for all functions in $\cF'$, then it must correctly learn all the $b_i$'s with probability at least $5/6$. 

\paragraph{Step 2: Learning $b_j$'s is equivalent to finding a point in $\cX_j$.} An algorithm can estimate $b_j$'s only through issuing appropriate queries to the oracle $\sO'$. For all $i \in \{1,2,\dots, d\}$, we define $\cX_i$ as
\begin{align}
    \cX_i := \left\{ x \in \cX \ \bigg| \ \left(\bigcap_{j < i} \{f_i(x) > f_j(x)\}\right) \cap \left(\bigcap_{j \geq i} \{f_i(x) \geq f_j(x)\}\right) \cap \left\{ |a_i^{\top} x| \leq \frac{1}{\sqrt{d}} \right\} \right\}.
\end{align}
Note that whenever an algorithm queries a point $x \in \cX_i$, the oracle returns $s_i(x) = -b_i$. Moreover, if the queried point $x \notin \cX_i$, the value returned by the oracle is either $-b_j$ for $j \neq i$ (when $x$ does not satisfy one of the first two conditions of being in $\cX_i$) or a fixed value in $\{-1, +1\}$ given by $\sgn(a_i^{\top}x)$, which is independent of the value of $b_i$ (when $x$ does not satisfy the third condition of being in $\cX_i$). Thus, in both cases, the output of the oracle is uncorrelated with $b_i$. Hence, querying a point $x \notin \cX_i$ yields no information about $b_i$. Consequently, an algorithm can learn $b_i$ \emph{only if} it can determine a point $x \in \cX_i$. Moreover, since $s_i(x)$ is noiseless, it is also sufficient to determine such an $x \in \cX_i$.

Hence, in order for an algorithm to estimate $b_i$, it needs to build an estimator for a point $x \in \cX_i$. Let $\varphi$ be an estimator for determining a point $x \in \cX_i$ such that $\Pr(\varphi \in \cX_i) < 2/3$. Here, we slightly abuse notation and use $\varphi$ to also denote the output of estimator $\varphi$. If $\hat{b}_i(\varphi)$ is an estimator of $b_i$ that uses $\varphi$, then $\sup_{\bfb} \Pr(\hat{b}_i(\varphi) \neq b_i) \geq \sup_{\bfb} \Pr(\hat{b}_i(\varphi) \neq b_i|\varphi \notin \cX_i)\Pr(\varphi \notin \cX_i) > \frac{1}{2}\cdot\frac{1}{3} = \frac{1}{6}$. Here, we used the observation that the output of the oracle at a point $x \notin \cX_i$ is uncorrelated with $b_i$ and hence cannot be better than a random guess. Thus, to correctly determine all $b_i$'s with probability $5/6$, an algorithm $\sA$ needs to determine a set of points $(\tilde{x}_1, \tilde{x}_2, \dots, \tilde{x}_d)$ such that 
\begin{align}
    \Pr \left( \bigcap_{i = 1}^d \{ \tilde{x}_i \in \cX_i\} \right) \geq \frac{2}{3}. \label{eqn:tilde_x_in_X_i_condition}
\end{align}

\paragraph{Step 3: Hardness of finding a point in $\cX_i$.} In this step, we characterize the hardness of finding a point $w \in \cX_i$ for some fixed $i \in \{1,2,\dots, d\}$. To characterize the hardness, we make use of the reduction outlined in the following lemma, whose proof is deferred to Appendix~\ref{proof_lemma:X_i_implies_X_prime_i}.
\begin{lemma} 
Let     
    \begin{align}
        \cX_i' := \left\{ x \in \cX \bigg| \left(\bigcap_{j < i} \left\{ \ip{a_i b_i}{x} < \ip{a_j b_j}{x} \right\} \right) \cap \left(\bigcap_{j \geq i} \left\{ \ip{a_i b_i}{x} \leq \ip{a_j b_j}{x} \right\} \right) \cap \left\{ |\ip{a_i b_i}{x}| \leq \frac{1}{\sqrt{d}} \right\} \right\} \label{eqn:X_prime_i_definition}
    \end{align}
    for all $i \in [d]$. It follows $
        \{w \in \cX_i\} \implies \{w \in \cX_i'\}$.
    \label{lemma:X_i_implies_X_prime_i}
\end{lemma}
In other words, finding a point $w \in \cX_i$ is at least as hard as finding a point $w \in \cX_i'$. Consequently, any set of points $\{\tilde{x}_1, \tilde{x}_2,\dots, \tilde{x}_d\}$
that satisfy~\eqref{eqn:tilde_x_in_X_i_condition} must also satisfy
\begin{align}
    \Pr \left( \bigcap_{i = 1}^d \{ \tilde{x}_i \in \cX_i'\} \right) \geq \frac{2}{3}. \label{eqn:tilde_x_in_X_prime_i_condition}
\end{align}
Moreover, note that by definition, the sets $\{\cX_i'\}_{i = 1}^d$ are disjoint and thus the points $\{\tilde{x}_1, \tilde{x}_2,\dots, \tilde{x}_d\}$ need to be distinct.

For the rest of the proof, we focus on characterizing the hardness of finding a set of points $\{\tilde{x}_1, \tilde{x}_2,\dots, \tilde{x}_d\}$ satisfying Eqn.~\eqref{eqn:tilde_x_in_X_prime_i_condition}. We claim that any routine $\sM$ that determines a set of points satisfying Eqn.~\eqref{eqn:tilde_x_in_X_prime_i_condition} can also determine a set of vectors $(y_1, y_2, \dots, y_{d'})$ such that $\ip{v_j}{y_j} > 0$ holds for all $j \leq d'$ with probability $2/3$ for some $(d-1)/2 \leq d' \leq d$ and $\{v_1, v_2, \dots, v_{d'}\} \subseteq \{a_1 b_1, a_2b_2, \dots, a_d b_d\}$. To prove this claim, we define
\begin{align}
    \cX_{\text{sol}} := \left\{ (w_1, w_2, \dots, w_d) \in \cX^d \ | \ w_j \in \cX_j' \quad \forall \ j \in \{1,2,\dots, d\} \right\}
\end{align}
to be the set of all possible solutions. For all $(w_1, w_2, \dots, w_d) \in \cX_{\text{sol}}$ define,
\begin{align}
    n_{+}(w_1, w_2, \dots, w_d) := |\{ j \ | \ \ip{a_j b_j}{w_j} \geq 0\}|; \quad n_{-}(w_1, w_2, \dots, w_d) := |\{ j \ | \ \ip{a_j b_j}{w_j} < 0\}|.
\end{align}
Lastly,  let
\begin{align}
    \cX_{\text{sol}}^{+} & := \left\{ (w_1, w_2, \dots, w_d) \in \cX_{\text{sol}} \ | \ n_{+}(w_1, w_2, \dots, w_d) > d/2 \right\} , \\
    \cX_{\text{sol}}^{-} & := \left\{ (w_1, w_2, \dots, w_d) \in \cX_{\text{sol}} \ | \ n_{-}(w_1, w_2, \dots, w_d) \geq d/2 \right\}.
\end{align}
It is straightforward to note that $\cX_{\text{sol}}^{+} $ and $ \cX_{\text{sol}}^{-} $ form a partition of $\cX_{\text{sol}}$. Thus, $\sM$ can determine an element of $\cX_{\text{sol}}$ with probability $2/3$ \emph{only if} it can either find an element in $\cX_{\text{sol}}^{+} $ with probability $2/3$ or find an element in $\cX_{\text{sol}}^{-} $ with probability $2/3$. 
Let us consider the two possible cases.
\begin{itemize}
    \item \textbf{Case (i): $\sM$ finds a point in $\cX_{\text{sol}}^{-} $.} By definition of $\cX_{\text{sol}}^{-}$, we know that there exists a set of indices $\{j_1, j_2, \dots, j_{d'}\}$ with $d' \geq d/2$ for which $\ip{a_{j_r}b_{j_r}}{\tilde{x}_{j_r}} < 0$ holds for all $r \in [d']$. If we choose $\{v_1, \dots, v_d\}$ such that $v_r = a_{j_r}b_{j_r}$, then $(y_1, y_2, \dots, y_{d'}) = (-\tilde{x}_{j_1}, -\tilde{x}_{j_2}, \dots, -\tilde{x}_{j_{d'}})$ is the required set of vectors. Note that finding a point $\tilde{x}$ is statistically equivalent to finding a point $-\tilde{x}$.
    \item \textbf{Case (ii): $\sM$ finds a point in $\cX_{\text{sol}}^{+} $.} By definition of $\cX_{\text{sol}}^{+}$, we know that there exists a set of indices $\{j_1, j_2, \dots, j_{d''}\}$ with $d'' > d/2$ for which $\ip{a_{j_r}b_{j_r}}{\tilde{x}_{j_r}} \geq 0$ holds for all $r \in [d'']$. WLOG, we assume that $j_1 < j_2 < \dots < j_{d''}$. Note that from the definition of $\cX_i'$, we can conclude that $\ip{a_{j_r}b_{j_r}}{\tilde{x}_{j_{r-1}}} > 0$ holds for all $r = 2,3,\dots, d''$. Thus, $\sM$ determines the required collection $(y_1, y_2, \dots, y_{d'}) = (\tilde{x}_{j_1}, \tilde{x}_{j_2}, \dots, \tilde{x}_{j_{d''-1}})$ corresponding to the subset $\{v_1, v_2, \dots, v_{d'}\} = \{a_{j_2}b_{j_2}, a_{j_3}b_{j_3}, \dots, a_{j_{d''}}b_{j_{d''}}\}$ with $d' = d'' - 1 > d/2 - 1$.
\end{itemize}

On combining the cases, we arrive at the statement. Consequently, we can conclude that determining a solution $\{\tilde{x}_1, \tilde{x}_2,\dots, \tilde{x}_d\}$ satisfying~\eqref{eqn:tilde_x_in_X_prime_i_condition} is at least as hard as finding a set of points $(y_1, y_2, \dots, y_{d'})$ such that $\ip{v_j}{y_j} > 0$ holds for all $j \leq d'$ with probability $2/3$ for some $(d-1)/2 \leq d' \leq d$ and $\{v_1, v_2, \dots, v_{d'}\} \subseteq \{a_1 b_1, a_2b_2, \dots, a_d b_d\}$.

\paragraph{Step 4: Establishing the equivalence to mean estimation.} Consider the problem of estimating a vector $y$ such that $\ip{\mu}{y} > 0$ w.p. $2/3$, using samples of the form $\cN(\mu, (\sigma^2/d) I_d)$, where $\mu$ is an unknown vector. We claim that this problem is at least as hard as solving the Gaussian mean estimation problem to within an error of $C\|\mu\|_2^2$ for some numerical constant $C > 0$.

To establish this claim, note that under the aforementioned model, the problem is at least as hard as finding $\widehat{\mu}$, an estimate of $\mu$ based on the samples, satisfying $\ip{\mu}{\widehat{\mu}} > 0$ w.p. $2/3$. Let $Z = \mu - \widehat{\mu}$ denote the estimation error and $u = \mu/\|\mu\|_2$ denote a unit vector in the direction of $\mu$. Note that under this model, $Z$ is independent of $\mu$. We have, 
\begin{align*}
    \ip{\mu}{\widehat{\mu}} > 0 \implies \ip{\mu}{\widehat{\mu} - \mu} > -\|\mu\|^2_2 \implies \ip{u}{Z} < \|\mu\|.
\end{align*}
Consequently, any estimator $\widehat{\mu}$ that ensures $\ip{\mu}{\widehat{\mu}} > 0$ w.p. $2/3$ for all choices of $u$ must ensure $\sup_{u} \ip{u}{Z} < \|\mu\|$ holds w.p. $2/3$, or equivalently, $\|Z\| \leq \|\mu\|$ with probability at least $2/3$. This is equivalent to solving the Gaussian mean estimation problem such that $\|\hat{\mu} - \mu\|^2 \leq \|\mu\|^2$ with probability at least $2/3$.  

Note that the problem faced by the learner, i.e., of identifying the set of vectors $(y_1, y_2, \dots, y_{d'})$ corresponding to $\{v_1, v_2, \dots, v_{d'}\} \subseteq \{a_1 b_1, a_2b_2, \dots, a_d b_d\}$, is identical to the one outlined above. In particular, for $y_j$, $\mu = \alpha v_j$, and the samples correspond to the queries to the oracle. As a result, the problem of identifying the set of vectors $(y_1, y_2, \dots, y_{d'})$ is equivalent to solving $d' = \Theta(d)$ mean estimation problems to within an error proportional to $\alpha$, the norm of the mean vector.

\subsection{Establishing the final bound} 

We can restate the problem of interest as follows. Let $\{\theta_1, \dots, \theta_{d'}\}$ be a collection of distinct vectors with (unknown) norm $\alpha$. Let $\sA'_{ME}$ be any distributed mean estimation algorithm with $M$ clients such that whose communication cost matches that of our optimization algorithm $\sA$, i.e.,  $\CC(\sA'_{ME}) = \CC(\sA)$. For any vector $\theta_j$, each client can query the oracle to obtain an independent sample from  $\cN(\theta_j, (\sigma^2/d)I_d)$. Using a total of $N$ such samples at each agent, the algorithms need to determine a set of estimates $\{\hat{\theta}_1, \dots, \hat{\theta}_{d'}\}$ such that with probability at least $2/3$,
\begin{align}
    \max_{j \leq d'} \|\hat{\theta}_j - \theta_j\|^2 \leq  \|\theta_1\|_2^2 =  
 \alpha^2. \label{eqn:theta_final_estimation_condition}
\end{align}

For all $j$, let $N_j$ and $B_j$ denote the number of samples used and the number of bits transmitted by each client respectively to estimate $\theta_j$.\footnote{For simplicity of exposition, we assume that the values $N_j$ and $B_j$ are same across all clients. This idea can be extended to the general case with different values at different clients using the sequence of arguments outlined in~\cite{Duchi2014DistributedEstimation}.} Using the results for distributed mean estimation~\citep{Braverman2016CommunicationLowerBounds, Barnes2020LowerBoundComm, Duchi2014DistributedEstimation}, we can conclude that 
\begin{align}
    \|\hat{\theta}_j - \theta_j\|^2  \geq c_0 \cdot \min \left\{  \frac{\sigma^2 d}{d N_j}, \max \left\{ \frac{\sigma^2 d^2}{d MN_jB_j}, \frac{\sigma^2 d}{d MN_j} \right\} \right\}  \geq c_0 \cdot \min \left\{  \frac{\sigma^2}{ N_j}, \max \left\{ \frac{\sigma^2 d}{MN_jB_j}, \frac{\sigma^2 }{ MN_j} \right\} \right\},  \label{eqn:max_estimation_error}
\end{align}
holds for each $j \leq d'$ w.p. at least $1/3$ where $c_0 > 0$ is a numerical constant. 

Define $\cJ_1 := \{j : N_j > 3N/d'\}$ and $\cJ_2 := \{j : B_j > 3\CC(\sA'_{ME})/d'\}$. It is straightforward to note that $|\cJ_1| \leq d'/3$ and $|\cJ_2| \leq d'/3$. Consequently, $|\cJ_1^c \cap \cJ_2^c| = d' - |\cJ_1 \cup \cJ_2| \geq d' - |\cJ_1| - |\cJ_2| \geq d'/3 > 0$. This implies that there exists an index $j'$ such that $N_{j'} \leq 3N/d'$ and $B_{j'} \leq 3\CC(\sA'_{ME})/d'\}$. Using this choice of $j'$ along with~\eqref{eqn:max_estimation_error} and the relations $d' \geq (d - 1)/2$ and $\CC(\sA'_{ME}) = \CC(\sA)$, we can conclude that
\begin{align}
    \max_{j \leq d'} \|\hat{\theta}_j - \theta_j\|^2 \geq  c_1 \cdot \min \left\{  \frac{\sigma^2d}{ N}, \max \left\{ \frac{\sigma^2 d^3}{MN \CC(\sA)}, \frac{\sigma^2 d}{ MN} \right\} \right\} \label{eqn:theta_est_err_final_lower_bound}
\end{align}
holds with probability at least $1/3$ for some numerical constant $c_1 > 0$.

Let us consider the scenario where 
\begin{align}
    \alpha^2 := \min\left\{\frac{c_1}{2} \cdot \min \left\{  \frac{\sigma^2d}{ N}, \max \left\{ \frac{\sigma^2 d^3}{MN \CC(\sA)}, \frac{\sigma^2 d}{ MN} \right\} \right\}, 1 \right\}. \label{eqn:optimal_alpha}
\end{align}
For this choice of $\alpha$, based on Eqn.~\eqref{eqn:theta_est_err_final_lower_bound} we can conclude that $\sA$ cannot solve the mean estimation problems to the required level of precision. As elaborated in the previous step, this implies that $\sA$ cannot identify the set of points $\{\tilde{x}_1, \tilde{x}_2, \dots, \tilde{x}_d\}$ with the required confidence and hence
\begin{align}
    \ER(\sA) = \sup_{f \in \cF} \ER(\sA; f) & \geq \sup_{f \in \cF'} \ER(\sA; f) \nonumber \\
    & \geq \frac{\alpha}{6\sqrt{d}}   \gtrsim \min\left\{  \min \left\{  \sqrt{\frac{\sigma^2}{ N}}, \max \left\{ \sqrt{\frac{\sigma^2 d^2}{MN \CC(\sA)}}, \sqrt{\frac{\sigma^2 }{ MN}} \right\} \right\}, \frac{1}{\sqrt{d}} \right\}. \label{eqn:sco_communication_lower_bound_r_2}
\end{align} 
As mentioned at the beginning of the proof, the above analysis can be easily extended to a domain with diameter $R$ by replacing $b_i$ with $Rb_i/2$ in the definition of the functions $f_i$. In a such a case, the corresponding relation for Eqn.~\eqref{eqn:low_error_implies_learning_b} would read as
\begin{align}
    \sup_{f \in \cF'} \E[\ER(\sA, f)] \leq \frac{R\alpha}{12\sqrt{d}} \implies \max_{j} \sup_{(A,\bfb)} p_j(\sA; A,\bfb) \geq \frac{5}{6}.
\end{align}
Consequently, Eqn.~\eqref{eqn:sco_communication_lower_bound} would be updated as
\begin{align}
    \ER(\sA) = & \geq \frac{R\alpha}{12\sqrt{d}}   \gtrsim R \min\left\{  \min \left\{  \sqrt{\frac{\sigma^2}{ N}}, \max \left\{ \sqrt{\frac{\sigma^2 d^2}{MN \CC(\sA)}}, \sqrt{\frac{\sigma^2 }{ MN}} \right\} \right\}, \frac{1}{\sqrt{d}} \right\}. \label{eqn:sco_communication_lower_bound}
\end{align}
The statistical term and the privacy term in the statement of Theorem~\ref{thm:lower_bound} follow from the standard lower bounds established in the literature~\citep{Agarwal2009SCOLowerBound, Bassily2014PrivateERMLowerBounds, Levy2021UserLevelPrivacy}. Specifically, Theorem 1 from~\cite{Agarwal2009SCOLowerBound} states that the error rate of any convex optimization algorithm $\sA$ with a total of $MN$ queries to a (sub)gradient oracle corresponding to a $1$-Lipschitz function satisfies 
\begin{align}
    \ER(\sA) \geq c_2 \min\left\{ R\cdot \sqrt{\frac{\sigma^2}{MN}}, \frac{R}{\sqrt{d}}\right\}. \label{eqn:sco_statistical_lower_bound}
\end{align}
To obtain the lower bound corresponding to the private estimation, note that the problem considered in this work is at least as hard as stochastic convex optimization in a centralized setting with $MN$ samples out of which $M$ samples can change in neighboring datasets (akin to user-level privacy). Thus, using the corresponding lower bounds from~\cite{Bassily2014PrivateERMLowerBounds, Levy2021UserLevelPrivacy}, we can conclude that 
\begin{align}
    \ER(\sA) \geq c_3 \cdot \frac{R\sqrt{d}}{\sqrt{M} N\varepsilon_{\DP}}. \label{eqn:sco_privacy_lower_bound}
\end{align}

On combining the results in Eqn.~\eqref{eqn:sco_communication_lower_bound},~\eqref{eqn:sco_statistical_lower_bound} and~\eqref{eqn:sco_privacy_lower_bound}, we arrive at the final result. \\

\paragraph{Extension to $L$-Lipschitz functions.} The above analysis can be easily extended to accommodate $L$-Lipschitz functions. In particular, for the hard instance, we replace $f(x)$ with $Lf(x)$ and carry out the same series of steps. As a result, Eqn.~\eqref{eqn:low_error_implies_learning_b} gets modified to
\begin{align}
    \sup_{f \in \cF'} \E[\ER(\sA, f)] \leq \frac{RL\alpha}{12\sqrt{d}} \implies \max_{j} \sup_{(A,\bfb)} p_j(\sA; A,\bfb) \geq \frac{5}{6}. \label{eqn:low_error_implies_learning_b_L_lipschitz}
\end{align}
Moreover, since the gradients scale by a factor of $L$, the condition in Eqn.~\eqref{eqn:theta_final_estimation_condition} changes to
\begin{align}
    \max_{j \leq d'} \|\hat{\theta}_j - \theta_j\|^2 \leq  \|\theta_1\|_2^2 =  
 L^2\alpha^2
\end{align} 
and the corresponding choice of $\alpha$ needs to be updated to 
\begin{align}
    \alpha^2 := \frac{1}{L^2}\min\left\{\frac{c_1}{2} \cdot \min \left\{  \frac{\sigma^2d}{ N}, \max \left\{ \frac{\sigma^2 d^3}{MN \CC(\sA)}, \frac{\sigma^2 d}{ MN} \right\} \right\}, 1 \right\}. 
\end{align}
In the light of Eqn.~\eqref{eqn:low_error_implies_learning_b_L_lipschitz}, the above choice of $\alpha$ results in a lower bound identical to Eqn.~\eqref{eqn:sco_communication_lower_bound}. Due to a similar flavour of analysis, the relation in Eqn.~\eqref{eqn:sco_statistical_lower_bound} also does not get affected by the choice of $L$. However, the change in lipschitz constant results in a change of sensitivity in the gradient estimation procedures. As a result, Eqn.~\eqref{eqn:sco_privacy_lower_bound} exhibits a linear dependence with $L$ for $L$-Lipschitz functions.

\subsection{Proof of Lemma~\ref{lemma:X_i_implies_X_prime_i}}
\label{proof_lemma:X_i_implies_X_prime_i}
Throughout the proof, we fix a vector $w \in \cX_i$. For simplicity of presentation, we use the shorthand 
\begin{align*}
    \alpha_j := \ip{a_j b_j}{w}
\end{align*}
for all $j \in [d]$. Using this shorthand, we can rewrite the condition $w \in \cX_i$ as
\begin{align*}
    \{w \in \cX_i\} = \left(\bigcap_{j < i} \left\{ \left| \frac{1}{\sqrt{d}} - \alpha_i \right| > \left| \frac{1}{\sqrt{d}} - \alpha_j \right| \right\} \right) \cap \left(\bigcap_{j \geq i} \left\{ \left| \frac{1}{\sqrt{d}} - \alpha_i \right| \geq \left| \frac{1}{\sqrt{d}} - \alpha_j \right| \right\} \right) \cap \left\{ |\alpha_i| \leq \frac{1}{\sqrt{d}} \right\}.
\end{align*}
To establish the statement of the lemma, we fix a value of $j$ and define the following events: 
\begin{subequations}
\begin{align} 
    \cE_0 & := \cE_1 \cap \cE_2 ,\label{eqn:e_0_def} \\
    \cE_1 & := \left\{\left| \frac{1}{\sqrt{d}} - \alpha_i \right| > \left| \frac{1}{\sqrt{d}} - \alpha_j \right|\right\}, \\
    \cE_2 & := \left\{ |\alpha_i| \leq \frac{1}{\sqrt{d}} \right\} ,\\
    \cE_3 & := \{ |\alpha_j| < |\alpha_i| \},\\
    \cE_4 & := \{\alpha_i < 0\}, \\
    \cE_5 & := \{\alpha_j > 0\}.
\end{align}
\end{subequations}

Let us first analyze the condition for $j < i$. Given $\cE_2 \cap \cE_3 \cap \cE_4^c$, we have,
\begin{align}
    \left| \frac{1}{\sqrt{d}} - \alpha_i \right| \overset{(a)}{=} \frac{1}{\sqrt{d}} - \alpha_i \overset{(b)}{=} \frac{1}{\sqrt{d}} - |\alpha_i| \overset{(c)}{\leq}  \frac{1}{\sqrt{d}} - |\alpha_j| \leq \left| \frac{1}{\sqrt{d}} - |\alpha_j| \right| \overset{(d)}{\leq} \left| \frac{1}{\sqrt{d}} - \alpha_j \right|,
\end{align}
where $(a), (b)$ and $(c)$ are a consequence of $\cE_2, \cE_4^c$ and $\cE_3$ respectively, and $(d)$ follows from triangle inequality. As a result, $\cE_1 \cap \cE_2 \cap \cE_3 \cap \cE_4^c = \emptyset$ which implies $\cE_1 \cap \cE_2 \cap \cE_3 = \cE_1 \cap \cE_2 \cap \cE_3 \cap \cE_4$. Similarly, given $\cE_2 \cap \cE_3^c \cap \cE_5^c$, we have,
\begin{align}
    \left| \frac{1}{\sqrt{d}} - \alpha_i \right| \overset{(a)}{=} \frac{1}{\sqrt{d}} - \alpha_i \leq \frac{1}{\sqrt{d}} + |\alpha_i| \overset{(b)}{\leq}  \frac{1}{\sqrt{d}} + |\alpha_j|  \overset{(c)}{\leq} \left| \frac{1}{\sqrt{d}} - \alpha_j \right|,
\end{align}
where $(a), (b)$ and $(c)$ are a consequence of $\cE_2, \cE_3^c$ and $\cE_5^c$ respectively. As a result, $\cE_1 \cap \cE_2 \cap \cE_3^c \cap \cE_5^c = \emptyset$ and hence, $\cE_1 \cap \cE_2 \cap \cE_3^c = \cE_1 \cap \cE_2 \cap \cE_3^c \cap \cE_5$. Using these two relations, we can conclude that
\begin{align*}
    \cE_0 & = \cE_1 \cap \cE_2 \\
    & = (\cE_1 \cap \cE_2 \cap \cE_3) \cup (\cE_1 \cap \cE_2 \cap \cE_3^c) \\
    & = (\cE_1 \cap \cE_2 \cap \cE_3 \cap \cE_4) \cup (\cE_1 \cap \cE_2 \cap \cE_3^c \cap \cE_5) \\
    & \subseteq \cE_2 \cap ((\cE_3 \cap \cE_4) \cup (\cE_3^c \cap \cE_5)) \cap \{\alpha_i \neq \alpha_j\} \\
    & \subseteq \cE_2 \cap \{\alpha_i < \alpha_j\},
\end{align*}
where in the fourth step the condition $\{\alpha_i \neq \alpha_j\}$ is a consequence of $\cE_1$ and the last step follows by noting $((\cE_3 \cap \cE_4) \cup (\cE_3^c \cap \cE_5)) \cap \{\alpha_i \neq \alpha_j\} = \{\alpha_i < \alpha_j\}$. By a very similar sequence of arguments, we can also show that for all $j > i$
\begin{align*}
    \left\{\left| \frac{1}{\sqrt{d}} - \alpha_i \right| \geq \left| \frac{1}{\sqrt{d}} - \alpha_j \right|\right\} \cap \left\{ |\alpha_i| \leq \frac{1}{\sqrt{d}} \right\} \subseteq \left\{ |\alpha_i| \leq \frac{1}{\sqrt{d}} \right\} \cap \{\alpha_i \leq \alpha_j\}. 
\end{align*}
Note that the only difference in this case is we allow $\alpha_i = \alpha_j$. On combining the two cases, we can conclude that $w \in \cX_i \implies w \in \cX_i'$, where $\cX_i'$ is defined in Eqn.~\eqref{eqn:X_prime_i_definition}.

%% file: proof_upper_bound.tex
\section{Proof of Theorem~\ref{thm:upper_bound}}
\label{sec:proof_achievability}

We separately establish the accuracy, privacy and communication complexity guarantees of \charter. 

\paragraph{Communication Cost.} The bound on communication cost is straightforward. Note that in the learning stage, \charter \  quantizes each gradient such that it can be expressed in $d\cdot J_0$ bits ($J_0$ bits for each coordinate). Since each agent transmits $K$ such gradients, one for each of the $K$ iterations, the communication cost during the learning phase is $Kd J_0$ bits. During the verification phase, each client transmits $K+1$ scalars, where each scalar is expressed using $J_1$ bits. Thus, the communication cost during the verification stage is $(K+1)J_1$. On combining the two and plugging in the values from Section~\ref{sub:parameter_setting}, we obtain,
\begin{align*}
    \CC(\charter) & = KdJ_0 + (K+1)J_1 \nonumber \\
    & \leq C_1 \cdot \left(d^2 \log(dMN) \cdot \log \left(\frac{2D_0 N \varepsilon_{\DP}}{\sqrt{d} + \varepsilon_{\DP}\sqrt{N}} \right) + d \log(dMN) \cdot \log \left(\frac{2D_1 N \varepsilon_{\DP}}{\sqrt{d} + \varepsilon_{\DP}\sqrt{N}} \right)\right) = \tilde{\cO}(d^2),
\end{align*}
as required. Here $C_1 > 0$ is a numerical constant.

\paragraph{Privacy.} To establish the privacy guarantees, note that it is sufficient to establish that both stages of the algorithm are $(\varepsilon_{\DP}, \delta_{\DP})$ differentially private as they use distinct subsets of $\cD$. Since the analysis is identical for all the clients, we drop the subscript $m$ for notational simplicity. We begin with stating some useful lemmas followed by the proof. 

\begin{definition}\label{def_sensitivity}
    Let $f : \cZ^N \to \R^k$. The $\ell_2$ sensitivity of $f$ is defined as
    \begin{align*}
        \Delta_{2, f} := \sup_{\cD, \cD'} \|f(\cD) - f(\cD')\|_2,
    \end{align*}
    where $\cD, \cD' \subset \cZ^N$ are neighboring datasets.
\end{definition}

\begin{lemma}[Gaussian Mechanism~\citep{Dwork2006DPOGPaper}]
    Let $f : \cZ^N \to \R^k$ obeying Definition~\ref{def_sensitivity} and $Y \in \R^k$ be a random vector each of whose entries is an i.i.d. random variable drawn according to zero mean Gaussian with variance $\frac{2\log(5/(4\delta))\Delta_{2,f}^2}{\varepsilon^2}$. The algorithm 
    \begin{align*}
        \cA(\cD) = f(\cD) + Y
    \end{align*}
    is $(\varepsilon, \delta)$-differentially private.
    \label{lemma:gaussian_mechanism}
\end{lemma}

\begin{lemma}[Amplification by subsampling~\citep{Balle2018PrivacyAmplification}] 
    For $\varepsilon, \delta \in (0,1)$, let $\cA: \cZ^k \to \Theta$ be an $(\varepsilon, \delta)$-differentially private algorithm. For $N > k$ and a dataset $S \subset \cZ^N$, let $S^{\textsc{WOR}}$ be a dataset of size $k$ obtained by randomly sampling points from $S$ without replacement. Then $\cA'$ obtained via $\cA'(S) = \cA(S^{\textsc{WOR}})$ is a $\left( (e-1)\frac{k\varepsilon}{N}, \frac{k\delta}{N}\right)$-differentially private algorithm.
    \label{lemma:privacy_amplification_by_subsampling}
\end{lemma}

\begin{lemma}[Advanced Composition Theorem~\citep{Dwork2015AdaptiveComposition, Kairouz2015Composition}]
    For any $\varepsilon > 0, \delta \in [0,1]$ and $\tilde{\delta} \in [0,1]$, the class of $(\varepsilon, \delta)$-differentially private mechanisms satisfies $(\tilde{\varepsilon}_{\tilde{\delta}}, 1 - (1-\delta)^k(1 - \tilde{\delta}))$-differentially privacy under $k$-fold adaptive composition for
    \begin{align*}
        \tilde{\varepsilon}_{\tilde{\delta}} := \min\left\{ k\varepsilon, \frac{k(e^{\varepsilon} - 1)\varepsilon}{(e^{\varepsilon} + 1)} + \varepsilon\sqrt{2k\log\left( \min\left\{ e + \frac{\sqrt{k\varepsilon^2}}{\tilde{\delta}}, \frac{1}{\tilde{\delta}}\right\}\right)} \right\}.
    \end{align*}
    \label{lemma:adaptive_composition}
\end{lemma}

We begin with the main proof starting with the privacy guarantees of the learning stage. Note that the $\ell_2$ sensitivity of $\widehat{\partial \cL}^{(\textsf{NonPriv,b})}$ is $6KG_0/N$. Thus, using the privacy guarantees of Gaussian Mechanism (Lemma~\ref{lemma:gaussian_mechanism}), we can conclude that for all iterations $k$,
$\widehat{\partial \cL}^{(\textsf{Priv,b})}(x_k)$ is $(\varepsilon_0, \delta_0)$ private with respect to $\cD^{(1, k)}$, where
\begin{align}
    \varepsilon_0 := \varepsilon_{\DP} \cdot \sqrt{\frac{K}{15\log(2.5/\delta_{\DP})}}; \quad \delta_0 := \frac{\delta_{\DP}}{2}.
\end{align}
Using Lemma~\ref{lemma:privacy_amplification_by_subsampling} and the condition $\varepsilon_{\DP} \leq \frac{1}{\sqrt{K}}$, we can conclude that $\widehat{\partial \cL}^{(\textsf{Priv,b})}(x_k)$ $(\varepsilon_1, \delta_1)$ private with respect to $\cD^{(1)}$, where
\begin{align}
    \varepsilon_1 := \frac{(e-1)\varepsilon_{\DP}}{2} \cdot \sqrt{\frac{1}{15K\log(2.5/\delta_{\DP})}}; \quad \delta_1 := \frac{\delta_{\DP}}{2K}.
\end{align}
Lastly, using Lemma~\ref{lemma:adaptive_composition} with $\tilde{\delta} = \delta_{\DP}/2$, we can conclude that \charter \  is $(\varepsilon_{\DP}, \delta_{\DP})$ differentially private during the learning stage. 

For the verification stage, note that for all $k$,  $\widehat{\cL}(x_k)$ is $(\varepsilon_2, \delta_2)$ private, where 
\begin{align}
    \varepsilon_2 := \varepsilon_{\DP} \cdot \sqrt{\frac{9}{20K\log(2.5/\delta_{\DP})}}; \quad \delta_2 := \frac{\delta_{\DP}}{2K}.
\end{align}
The final privacy guarantee of the verification stage then follows by again invoking Lemma~\ref{lemma:adaptive_composition} with $\tilde{\delta} = \delta_{\DP}/2$. 

\paragraph{Accuracy.} 
We establish the utility guarantees of \charter \ in four steps. In the first step, we establish that the loss estimates obtained at the end of the verification stage are close to the true values by bounding the estimation error during the verification stage. In the second step, we use these bounds to relate the excess risk of the algorithm to that of the minimum among the iterates. In the third step, we show that the iterates generated by the algorithm are such that there exists at least one iterate with sufficiently small excess risk. In the last step, we combine the results to obtain the final bound. \\

\noindent \textbf{Step $1$: Bounding the estimation error.} In the verification stage, we have the following relation for all $k \in  \{0,1, 2,\dots, K\}$
\begin{align}
  &  \widehat{\cL}(x_k) - \cL(x_k) \nonumber \\
  & = \underbrace{\frac{1}{M} \sum_{m = 1}^M (\widehat{\cL}_m(x_k) - \widehat{\cL}^{\textsf{Priv}}_m(x_k))}_{ := L_1} + \underbrace{\frac{1}{M} \sum_{m = 1}^M (\widehat{\cL}^{\textsf{Priv}}_m(x_k) - \widehat{\cL}^{\textsf{NonPriv}}_m(x_k))}_{:= L_2} + \underbrace{\frac{1}{M} \sum_{m = 1}^M (\widehat{\cL}^{\textsf{NonPriv}}_m(x_k) - \cL(x_k))}_{:= L_3}.
 \label{eqn:hat_l_verification}
\end{align}
We separately bound each of the three terms on the RHS.
\begin{itemize}
    \item \textbf{Bounding $L_1$:} We use the following lemma that provides concentration guarantees for clipped sub-Gaussian random variables to obtain a bound on $L_1$.
    \begin{lemma}{(Lemma B.1 from ~\cite{Salgia2023LinearBandits})}
        Let $X_1, X_2, \dots, X_n$ be a collection of i.i.d. $\sigma^2$-sub-Gaussian random variables with mean $\mu$. For all $i$, define $Y_i = X_i \1\{|x| \leq B\}$, where $B \geq |\mu| + \sigma\sqrt{2\log(4n)}$. Then, with probability $1 - \delta$,
        \begin{align*}
            \left| \frac{1}{n}\sum_{i = 1}^n Y_i - \mu \right| \leq \sigma \sqrt{\frac{2}{n}\log\left( \frac{4}{\delta}\right)}.
        \end{align*}
        \label{lemma:clipped_sub_gaussian_concentration}
    \end{lemma}
    Note that the prescribed choice of $G_1$ satisfies the condition in the above lemma. On invoking the above lemma along with the choice of $G_1$, we can conclude that 
    \begin{align}
        |L_1| \leq \sigma_f \sqrt{\frac{6}{MN} \log \left(\frac{32(K+1)}{\delta_{\Err}} \right)} \label{eqn:estimation_Err_conc_verification}
    \end{align}
    holds for $x_k$ with probability $1 - \delta_{\Err}/(8(K+1))$. Moreover, on taking a union bound over all $k$, we can conclude that the above relation holds for all $k$ with probability $1 - \delta_{\Err}/8$.
    \item \textbf{Bounding $L_2$:} The term $L_2$ corresponds to the error induced by the privatization noise. Since privatization just involves the addition of Gaussian noise, we can use the concentration of Gaussian random variables to bound $L_2$. Thus, 
    \begin{align}
        | L_2 | & \leq \sigma_{1} \sqrt{\frac{2}{M} \log \left( \frac{16(K+1)}{\delta_{\Err}}\right)}. \label{eqn:priv_Err_conc_verification} 
    \end{align}
    holds with probability $1 - \delta_{\Err}/8(K+1)$. Upon again invoking the union bound argument, we can conclude that the above relation holds for all $k$ with probability $1 - \delta_{\Err}/8$.
    \item \textbf{Bounding $L_3$:} On using the prescribed choice of $D_1$ along with the concentration of Gaussian random variables, we obtain that $|\widehat{\cL}^{\textsf{Priv}}_m(x_k))| \leq D_1$ holds for all $m, k$ with probability $1 - \delta_{\Err}/8$. Moreover, in the stochastic quantization routine, the quantization noise is bounded and hence sub-Gaussian with parameter $4D_1^2 \cdot 4^{-J_1}$. Consequently, the following relation holds for all $k$ with probability $1 - 2\delta_{\Err}/8$:
    \begin{align}
        | L_3| & \leq 2D_1 \cdot 2^{-J_1} \sqrt{\frac{2}{M} \log \left( \frac{16(K+1)}{\delta_{\Err}}\right)}. \label{eqn:quant_Err_conc_verification}
    \end{align}
\end{itemize}

On combining the relations in~\eqref{eqn:estimation_Err_conc_verification}, 
 and~\eqref{eqn:priv_Err_conc_verification},~\eqref{eqn:quant_Err_conc_verification}  and plugging them into~\eqref{eqn:hat_l_verification}, we obtain that
\begin{align}
  &  |\widehat{\cL}(x_k) - \cL(x_k)| \nonumber \\
  & \leq \sigma_f \sqrt{\frac{6}{MN} \log \left(\frac{32(K+1)}{\delta_{\Err}} \right)} + \sigma_{1} \sqrt{\frac{2}{M} \log \left( \frac{16(K+1)}{\delta_{\Err}}\right)} + 2D_1 \cdot 2^{-J_1} \sqrt{\frac{2}{M} \log \left( \frac{16(K+1)}{\delta_{\Err}}\right)} \label{eqn:verification_error}
\end{align}
holds with probability $1 - \delta_{\Err}/2$.  \\

\noindent \textbf{Step $2$: Relating the excess risks.} Let $k^{\star}$ be as defined in~\eqref{eqn:k_star_def} and $k^{\dagger}$ be
\begin{align}
    k^{\dagger} := \argmin_{k} \cL(x_k).
\end{align}
Then,
\begin{align}
    \cL(x_{k^{\star}}) \leq \widehat{\cL}(x_{k^{\star}}) + \zeta \leq \widehat{\cL}(x_{k^{\dagger}}) + \zeta \leq \cL(x_{k^{\dagger}}) + 2\zeta, \label{eqn:k_star_k_dagger_relation}
\end{align}
where $\zeta$ corresponds to the expression on the RHS in~\eqref{eqn:verification_error}.

This implies that the excess risk of the point output by the algorithm is at most an additive factor larger than that of the point with the smallest excess risk in $\{x_0, x_1, \dots, x_{K+1}\}$. Thus, it is sufficient for \charter \ to ensure that at least one iterate obtained during the learning stage has a small excess risk. We analyze the performance of the learning stage in the step to establish the existence of such a point.  \\

\noindent \textbf{Step $3$: Existence of an iterate with small excess risk.} Our analysis in this step builds upon the analysis of Vaidya's method~\citep{Vaidya1996PlaneCutting, Anstreicher1997Vaidya}. Let $x_c$ be the center of $\cX$ and $\cX_0$ be the set given by 
\begin{align}
    \cX_0 = \left\{x_c \pm \frac{R}{2\sqrt{d}} \cdot e_1, x_c \pm \frac{R}{2\sqrt{d}} \cdot e_2, \dots, x_c \pm \frac{R}{2\sqrt{d}} \cdot e_d\right\},
\end{align}
where $\{e_1, e_2, \dots, e_d\}$ denote the canonical basis of $\R^d$. In other words, $\cX_0$ denote the vertices of an $\ell_1$-ball of radius $\frac{R}{2\sqrt{d}}$, centered at $x_c$. Note that $\cX_0 \subset \cX$. Let $x^{\star}$ be any fixed minimizer of the function $f$ in $\cX$ and $\cX_1$ be the set given by
\begin{align}
    \cX_1 := (1 - \omega) x^{\star} + \omega \cX_0,
\end{align}
where $\omega = \frac{\sigma_{\max}}{\sqrt{MN}}$. Thus, $\text{conv}(\cX_1)$ is an $\ell_1$-ball of radius $\frac{R\sigma_{\max}}{2\sqrt{dMN}}$, centered at $x^{\star}+ \omega (x_c - x^{\star})$. Here $\text{conv}(\cY)$ denotes the convex hull of the set $\cY$. Using convexity of $\cX$ and the relation $\cX_0 \subset \cX$, we can conclude that $\cX_1 \subset \cX$. Let $\overline{x}_1 \in \cX_1$ and $\overline{x}_0$ be the corresponding point in $\cX_0$. Thus, 
\begin{align}
    \|\overline{x}_1 - x^{\star}\|_2 = \omega\|\overline{x}_0 - x^{\star}\|_2 \leq \omega R = \frac{R\sigma_{\max}}{\sqrt{MN}}. \label{eqn:dist_of_x1_from_x_star}
\end{align}

We claim that there exists an iteration $k' \in \{0,2,\dots, K\}$ such that $\cX_1 \subset P_{k'}$ and $\cX_1 \not\subset P_{k'+1}$, where $P_k$ denotes the polyhedron $(A_k, b_k)$ constructed during Vaidya's method. In other words, during the iteration $k'$, one of the points in $\cX_1$ is eliminated. We defer the proof of the claim to the end of the section.

We show that $x_{k'}$ is the required point that has a small excess risk. Firstly, note that a point is eliminated in Vaidya's method only when a constraint is added. Secondly, recall that in the $k^{\text{th}}$ iteration, we add the constraint $c_k^{\top}x \geq \beta_k$, where $c_k = -\widehat{\partial\cL}(x_k)$ and $\beta_k \leq c_k^{\top}x_k$. This implies all points eliminated during the $k^{\text{th}}$ iteration satisfy $c_k^{\top}x < \beta_k \leq c_k^{\top}x_k$. Thus, if a point $\overline{x} \in \cX_1$ is eliminated in iteration $k'$, then $\overline{x}$ satisfies 
\begin{align}
    \ip{-\widehat{\partial\cL}(x_{k'})}{\overline{x} - x_{k'}} < 0 \implies  \ip{\widehat{\partial\cL}(x_{k'})}{\overline{x} - x_{k'}} > 0. \label{eqn:elimination_condition}
\end{align}
Consequently, 
\begin{align}
    \cL(x_{k'}) & < \cL(x_{k'}) +  \ip{\widehat{\partial\cL}(x_{k'})}{\overline{x} - x_{k'}} \nonumber \\
    & < \cL(x_{k'}) +  \ip{\partial\cL(x_{k'})}{\overline{x} - x_{k'}} + \ip{\widehat{\partial\cL}(x_{k'}) - \partial\cL(x_{k'})}{\overline{x} - x_{k'}}\nonumber \\
    & < \cL(\overline{x})  + \ip{\widehat{\partial\cL}(x_{k'}) - \partial\cL(x_{k'})}{\overline{x} - x_{k'}} \nonumber\\
    & < \cL(x^{\star}) + \frac{R\sigma_{\max}}{\sqrt{MN}}  + \ip{\widehat{\partial\cL}(x_{k'}) - \partial\cL(x_{k'})}{\overline{x} - x_{k'}}, \label{eqn:min_iterate_bound}
\end{align}
where the first line follows from~\eqref{eqn:elimination_condition}, the third line from the convexity of $\cL$ and the fourth line from \eqref{eqn:dist_of_x1_from_x_star} and $1$-Lipschitzness of $\cL$ (Assumption~\ref{ass:pop_loss_convexity}). Thus, if the error $\ip{\widehat{\partial\cL}(x_{k'}) - \partial\cL(x_{k'})}{\overline{x} - x_{k'}}$ is small, the excess risk at $k'$ is also small.

To establish this result, we first state a relation that will be useful for the analysis. We claim that 
\begin{align}
    \frac{1}{4} \leq T_{k,m} \cdot \frac{3K}{N} \leq 1 \label{eqn:t_k_m_bound}
\end{align}
holds for all clients $m$ and iterations $k$ with probability $1- \delta_{\Err}/10$ as long as $N = \Omega(d \log (MK))$. We defer the proof of the claim to the end of the section. Moreover, we carry out the remainder of the analysis conditioned on this event.

We establish that $\ip{\widehat{\partial\cL}(x_{k}) - \partial\cL(x_{k})}{\overline{x} - x_{k}}$ is small for all iterations $k$ which immediately yields the bound for iteration $k'$. We use a similar modus operandi as used in Step $1$. Consider the $k^{\text{th}}$ iteration and any fixed $\overline{x} \in \cX_1$. We have, 
\begin{align}
    \ip{\widehat{\partial\cL}(x_{k}) - \partial\cL(x_{k})}{\overline{x} - x_{k}} & = \underbrace{  \frac{1}{M} \sum_{m = 1}^M \ip{\frac{N}{3KT_{k,m}} \widehat{\partial \cL}^{\textsf{NonPriv,b}}_m(x_k) - \partial\cL(x_{k})}{\overline{x} - x_{k}}}_{:=W_1} \nonumber \\
    & + \underbrace{ \frac{1}{M} \sum_{m = 1}^M \frac{N}{3KT_{k,m}} \cdot \ip{\widehat{\partial \cL}^{\textsf{Priv,b}}_m(x_k) - \widehat{\partial \cL}^{\textsf{NonPriv,b}}_m(x_k)}{\overline{x} - x_{k}}}_{:=W_2} \nonumber \\
    & + \underbrace{\frac{1}{M} \sum_{m = 1}^M \ip{\widehat{\partial \cL}_m(x_k) - \widehat{\partial \cL}^{\textsf{Priv,u}}_m(x_k)}{\overline{x} - x_{k}}}_{:=W_3}. \label{eqn:gradient_estimate_err_decomposition}
\end{align}
Similar to Step $1$, we separately bound each of the three terms on the RHS of Eq.~\eqref{eqn:gradient_estimate_err_decomposition}.
\begin{itemize}
    \item \textbf{Bounding $W_1$:} To bound $W_1$, note that 
    \begin{align*}
    \frac{N}{3KT_{k,m}} \widehat{\partial \cL}^{\textsf{NonPriv,b}}_m(x_k) = \frac{1}{T_{k,m}} \sum_{z \in \cD_{m}^{(1,k)}} \textsf{clip}(\partial\ell(x_k; z);  G_0) \cdot 
     \1\{z \notin \cup_{j = 1}^{k-1} \cD_m^{(1, j)}\},
    \end{align*}
    is an estimate of $\partial\cL(x_k)$ using $T_{k,m}$ independent (clipped) samples. If $\overline{v}$ denotes the unit vector along $\overline{x} - x_k$, then using the sub-Gaussianity of the samples (Assumption~\ref{ass:sub_Gaussian_noise}), we know that $\ip{\partial \ell(x; z)}{\overline{v}}$ is a sub-Gaussian random variable with parameter $\sigma_g^2/d$. Moreover, the choice of $G_0$ satisfies the condition in Lemma~\ref{lemma:clipped_sub_gaussian_concentration} for the random variable $\ip{\ell(x; z)}{\overline{v}}$. Thus, using Lemma~\ref{lemma:clipped_sub_gaussian_concentration}, we can conclude that 
    \begin{align}
        |W_1| & = \|\overline{x} - x_k\|_2 \cdot \left|  \frac{1}{M} \sum_{m = 1}^M \left\langle \frac{N}{3KT_{k,m}} \widehat{\partial \cL}^{\textsf{NonPriv,b}}_m(x_k) - \partial\cL(x_{k}), \, \overline{v} \right\rangle\right| \nonumber \\
        & \leq R \sigma_g \sqrt{\log\left(\frac{80d(K+1)}{\delta_{\Err}}\right)  \cdot \frac{2}{M^2}\sum_{m = 1}^M \frac{1}{T_{k,m}}} \nonumber \\
        & \leq R \sigma_g \sqrt{\frac{24K}{dMN}\cdot \log\left(\frac{80d(K+1)}{\delta_{\Err}}\right)}, \label{eqn:estimation_err_bound_learning}
    \end{align}
    holds with probability $1 - \delta_{\Err}/(20d(K+1))$. Here, the last line follows using~\eqref{eqn:t_k_m_bound}. Using a union bound we obtain that the above relation holds for all $k$ with probability $1 - \delta_{\Err}/(20d)$.
    \item \textbf{Bounding $W_2$:} Note that $\frac{N}{3KT_{k,m}} \cdot \ip{\widehat{\partial \cL}^{\textsf{Priv,b}}_m(x_k) - \widehat{\partial \cL}^{\textsf{NonPriv,b}}_m(x_k)}{\overline{x} - x_{k}}$ is a Gaussian random variable with variance $\left(\frac{N}{3KT_{k,m}}\right)^2R^2\sigma_0^2 \leq 16R^2\sigma_0^2$ where the inequality follows from the bound in Eqn.~\eqref{eqn:t_k_m_bound}. Consequently, for all $k$,
    \begin{align}
        |W_2| \leq 4R\sigma_0 \cdot \sqrt{\frac{2}{M} \log\left(\frac{40d(K+1)}{\delta_{\Err}} \right)} \label{eqn:priv_err_bound_learning}
    \end{align}
    holds with probability $1 - \delta_{\Err}/(20d)$.
    \item \textbf{Bounding $W_3$:} Lastly, to bound $W_3$, we use the same approach as used for $L_3$. For the choice $D_0$ and in light of Eqn.~\eqref{eqn:t_k_m_bound}, we can conclude that the event $\cE = \{ \|\widehat{\partial \cL}^{\textsf{Priv,u}}_m(x_k)\|_{\infty} \leq D_0 \ \ \forall m,k\}$ holds with probability $1 - \delta_{\Err}/10$. Since each coordinate of the quantized vector is an independent sub-Gaussian random variable with parameter $4D_0^2 \cdot 4^{-J_0}$, $\ip{\widehat{\partial \cL}_m(x_k) - \widehat{\partial \cL}^{\textsf{Priv,u}}_m(x_k)}{\overline{x} - x_k}$ is a sub-Gaussian random variable with parameter $4D_0^2 \cdot 4^{-J_0} \cdot \|\overline{x} - x_k\|^2$. Using the concentration of sub-Gaussian random variables and a union bound argument, we can conclude that, conditioned on $\cE$,
    \begin{align}
        |W_3| \leq 2D_0 \cdot 2^{-J_0} \cdot R \cdot \sqrt{\frac{2}{M} \log\left(\frac{40d(K+1)}{\delta_{\Err}} \right)} \label{eqn:quant_err_bound_learning}
    \end{align}
    holds for all $k$ with probability $1 - \delta_{\Err}/(20d)$. Here, we used the relation $\|\overline{x} - x_k\| \leq R$.
\end{itemize}

Thus, the relations~\eqref{eqn:gradient_estimate_err_decomposition},~\eqref{eqn:quant_err_bound_learning},~\eqref{eqn:priv_err_bound_learning}, and~\eqref{eqn:estimation_err_bound_learning} taken together along with a union bound over $\overline{x} \in \cX_1$ imply that
\begin{align}
    |\ip{\widehat{\partial\cL}(x_{k}) - \partial\cL(x_{k})}{\overline{x} - x_{k}}| & \leq 2D_0 \cdot 2^{-J_0} \cdot R \cdot \sqrt{\frac{2}{M} \log\left(\frac{40d(K+1)}{\delta_{\Err}} \right)} + 4R\sigma_0 \cdot \sqrt{\frac{2}{M} \log\left(\frac{40d(K+1)}{\delta_{\Err}} \right)} \nonumber \\
    & ~~~~~~~~~~~~~~~~~~~~~~~~~~~~~~~~~~~~~ + R \sigma_g \sqrt{\frac{24K}{dMN}\cdot \log\left(\frac{80d(K+1)}{\delta_{\Err}}\right)} \label{eqn:sub_optimality_loss}
\end{align}
holds for all $k \in \{0,1,\dots, K\}$, $m \in \{1,2,\dots, M\}$ and $\overline{x} \in \cX_1$ with probability $1 - 3\delta_{\Err}/10$. \\

\noindent \textbf{Step $4$: Putting it together.} On combining~\eqref{eqn:verification_error},~\eqref{eqn:k_star_k_dagger_relation},~\eqref{eqn:min_iterate_bound}, and~\eqref{eqn:sub_optimality_loss}, plugging in the prescribed parameter values from Section~\ref{sub:parameter_setting} and accounting for the conditioning on the $\cE$ and Eqn.~\eqref{eqn:t_k_m_bound} , we obtain that
\begin{align}
   & \cL(x_{k^{\star}}) - \cL(x^{\star}) \nonumber \\
   & \leq C_1(R\sigma_g + \sigma_f)\sqrt{\frac{\log N}{MN} \cdot \log\left( \frac{d^2\log (MN)}{\delta_{\Err}}\right)} + C_2R'\frac{\sqrt{d}\log(MN)}{N\varepsilon_{\DP}} \log\left(\frac{d\log (MN)}{\delta_{\DP}} \right) \sqrt{\log\left( \frac{d^2\log (MN)}{\delta_{\Err}}\right)}
\end{align}
holds with probability $1 - \delta_{\Err}$ for some constants $C_1, C_2$ that are independent of all problem parameters and $R' = R(1 + \sigma_{g}) + \sigma_{f}$.

\paragraph{Proving the claim~\eqref{eqn:t_k_m_bound}.} To establish this result, firstly note that we sample (uniformly at random) $N/3K$ points for $K$ rounds from a dataset of size $2N/3$. Thus, for all rounds, the number of previously seen data points are at most $N/3$, which is half the dataset. To lower bound the value of $T_{k,m}$, we obtain an upper bound on $\tilde{T}_{k,m} = \frac{N}{3K} - T_{k,m}$, i.e., the number of points in the set that have been seen previously by the algorithm. Using Hoeffding's inequality, which also holds for sampling without replacement~\citep{Hoeffding1994Concentration, Bardenet2015ConcentrationWithoutReplacement}, we can conclude that with probability $1- \delta_{\Err}/(10M(K+1))$,
\begin{align}
    \tilde{T}_{k,m} \leq \frac{N}{3K} \cdot \frac{3N_{k, m}}{2N} + \sqrt{\frac{N}{6K} \log\left( \frac{10M(K+1)}{\delta_{\Err}} \right)},
\end{align}
where $N_{k,m}$ denotes the number of samples that have been seen before iteration $k$ at client $m$. As shown above, $N_{k,m} \leq N/3$ for all $k,m$ with probability $1$. Thus, if $N \geq 24K\log\left( \frac{10M(K+1)}{\delta_{\Err}} \right)$, then,
\begin{align}
    \tilde{T}_{k,m} \leq \frac{N}{3K} \cdot \left(\frac{1}{2} + \sqrt{\frac{3K}{2N} \log\left( \frac{10M(K+1)}{\delta_{\Err}} \right)} \right) \leq \frac{N}{3K} \cdot \left(\frac{1}{2} + \sqrt{\frac{1}{16} } \right) \leq \frac{3}{4} \cdot \frac{N}{3K}.
 \end{align}
On taking union bound over $k$ and $m$, we can conclude that 
\begin{align}
    T_{k,m} \geq \frac{1}{4} \cdot \frac{N}{3K}
\end{align}
holds for all $k \in \{0,1,\dots,K\}$ and $m \in \{1,2,\dots, M\}$ with probability $1 - \delta_{\Err}/10$. The upper bound on $T_{k, m}$ follows directly by definition.

\paragraph{Proving that $\overline{x} \in \cX_1$ is eliminated.} We establish this claim using contradiction. Specifically, if we assume $\cX_1 \subset P_k$ for all $k \leq K$, then $P_k$ contains an $\ell_1$ ball of radius $\frac{\omega R}{2\sqrt{d}}$ for all $k \leq K$. This is because $P_k$ is a convex set and if $\cX_1 \subset P_k$, then the convex hull of $\cX_1$, which is an $\ell_1$ ball, also lies in $P_k$. If for all $k \leq K$, $P_k$ contains an $\ell_1$ ball of radius $\frac{\omega R}{2\sqrt{d}}$, then for all $k \leq K$
\begin{align}
    \log (\vol(P_k)) \geq d \log\left(\frac{\omega R}{\sqrt{d}}\right) - \log(d!) \geq d \log\left( \frac{R\sigma_{g}}{d\sqrt{dMN}}\right). \label{eqn:p_k_volume_lower_bound}
\end{align}
The RHS is the logarithm of the volume of an $\ell_1$ ball of radius $\frac{\omega R}{2\sqrt{d}}$, where $d!$ denotes the factorial of $d$. On the other hand,~\cite{Vaidya1996PlaneCutting} shows that the volume of the polyhedron after $k^{\text{th}}$ iteration of Vaidya's method is given by
\begin{align}
    \log(\vol(P_k)) \leq d \log \left(\frac{2d}{\gamma}\right) - V^0 - \frac{\gamma k}{2}, \label{eqn:p_k_volume}
\end{align}
where $\gamma$ is the parameter of Vaidya's algorithm and $V^0$ is the initial volumetric barrier. Since we start with a hypercube, its volumetric center is the same as the geometric center of the hypercube. Consequently, the volumetric barrier of a cube of side $b$ is given by 
\begin{align}
    V_{\text{cube}} = \frac{d}{2} \log \left( \frac{8}{b^2} \right).
\end{align}
Since the diameter of $\cX$ is $R$, the initial volumetric barrier is given by
\begin{align}
    V^{0} = \frac{d}{2} \log \left( \frac{8d}{R^2} \right).
\end{align}
On plugging the above relation into~\eqref{eqn:p_k_volume} along with the value of $K$, we obtain,
\begin{align*}
    \log(\vol(P_K)) & \leq d \log \left(\frac{2d}{\gamma}\right) - \frac{d}{2} \log \left( \frac{8d}{R^2} \right) - \frac{\gamma}{2} \cdot \frac{4d}{\gamma}\log\left(\frac{d\sqrt{MN}}{\gamma \sigma_{g}}\right) \\
    & \leq d \log \left(\frac{2d}{\gamma} \cdot \frac{R}{\sqrt{8d}} \cdot \frac{\gamma \sigma_{g}}{d^2MN}\right)  \\
    & \leq d \log \left(\frac{R\sigma_{g}}{dMN\sqrt{2d}}\right).
\end{align*}
This results in a contradiction with the lower bound on the volume of $P_K$ from~\eqref{eqn:p_k_volume_lower_bound}. This implies that $\cX_1 \not\subset P_K$ and hence some $\overline{x} \in \cX_1$ was eliminated during the algorithm.